\newif\ifOneColumn

\ifOneColumn
  \documentclass[journal,12pt,onecolumn,draftclsnofoot]{IEEEtran}
\else
  \documentclass[journal]{IEEEtran}
\fi
\usepackage{amsmath,amsfonts}
\usepackage{algorithmic}
\usepackage{algorithm}
\usepackage{array}
\usepackage[caption=false,font=normalsize,labelfont=sf,textfont=sf]{subfig}
\usepackage{textcomp}
\usepackage{stfloats}
\usepackage{url}
\usepackage{verbatim}
\usepackage{graphicx}
\usepackage{cite}
\usepackage[capitalize]{cleveref}
\usepackage{amsthm,multirow,enumitem,colortbl}
\usepackage[T1]{fontenc}
\definecolor{LightCyan}{rgb}{0.88,1,1}
\newcommand\Tstrut{\rule{0pt}{2.5ex}}     
\newcommand\Bstrut{\rule[-1.0ex]{0pt}{0pt}}
\newcommand{\TBstrut}{\Tstrut\Bstrut} 
\theoremstyle{plain}
\newtheorem{theorem}{Theorem}

\newtheorem{lemma}{Lemma}

\theoremstyle{definition}

\theoremstyle{remark}

\begin{document}

\title{Reinforcement Learning with \\ Non-Cumulative Objective}

\author{\IEEEauthorblockN{Wei Cui, \IEEEmembership{Student Member,~IEEE}, and Wei Yu, \IEEEmembership{Fellow,~IEEE}}   
\thanks{Manuscript submitted on November 10, 2022, revised on \today. This work is supported by Natural Sciences and Engineering Research Council (NSERC) of Canada via the Canada Research Chairs Program.  
The authors are with The
Edward S.~Rogers Sr.~Department of Electrical and Computer Engineering,
University of Toronto, Toronto, ON M5S 3G4, Canada 
(e-mails: \{cuiwei2, weiyu\}@ece.utoronto.ca).}}



\maketitle

\begin{abstract}
In reinforcement learning, the objective is almost always defined as a \emph{cumulative} function over the rewards along the process. However, there are many optimal control and reinforcement learning problems in various application fields, especially in communications and networking, where the objectives are not naturally expressed as summations of the rewards. In this paper, we recognize the prevalence of non-cumulative objectives in various problems, and propose a modification to existing algorithms for optimizing such objectives. Specifically, we dive into the fundamental building block for many optimal control and reinforcement learning algorithms: the Bellman optimality equation. To optimize a non-cumulative objective, we replace the original \emph{summation operation} in the Bellman update rule with a generalized operation corresponding to the objective. Furthermore, we provide sufficient conditions on the form of the generalized operation as well as assumptions on the Markov decision process under which the globally optimal convergence of the generalized Bellman updates can be guaranteed. We demonstrate the idea experimentally with the \emph{bottleneck} objective, i.e., the objectives determined by the \emph{minimum} reward along the process, on classical optimal control and reinforcement learning tasks, as well as on two network routing problems on maximizing the flow rates.
\end{abstract}


\section{Introduction}
In reinforcement learning (RL), an agent performs a sequence of actions to optimize a certain objective, over an environment modeled as a Markov decision process (MDP) \cite{MDP}. The objective value is determined by the collection of intermediate rewards the agent receives until the MDP is terminated (or an absorbing state is reached). In most of the literature, the objective is defined as the summation of these intermediate rewards, which corresponds to the \emph{summation operation} in the Bellman optimality equation \cite{bellman} when computing the value function. Such \emph{cumulative} objectives indeed capture the ultimate goals for many problems, such as Atari games \cite{dqn}, stock trading \cite{deep_hedging}, advertisement placements \cite{rl_advertise}, and so on. Nonetheless, there are many problems with objectives that do not translate to summations of rewards. 

Specifically, in the field of wireless communications, there are many system optimization problems that can be formulated and decomposed into sequences of optimization decisions, whose global objectives cannot be readily expressed as summations of rewards from individual optimization decisions. Examples of such problems include but are not limited to \emph{max-min} optimizations in routing and resources allocation \cite{drl_routing,haohang,manijeh,hailiang}, \emph{harmonic mean} maximization for traffic engineering \cite{mina} and for transmission system optimization \cite{mazen,mazen2}, the \emph{proportional fairness} optimizations for wireless communications \cite{giuseppe,proportional,cluster}, and so on. In this paper, we recognize the prevalence of problems with non-cumulative objectives, and propose modifications to many existing optimal control and RL algorithms for optimizing such objectives\footnote{The code for this paper is available at:  \url{https://github.com/willtop/Reinforcement_Learning_With_Non-Cumulative_Objective}}. 

In the optimal control or reinforcement learning literature, one class of problems with non-cumulative objectives are the problems where only terminal states matter, such as the the game of \emph{Go} \cite{alphaGo1,alphaGo2} or \emph{Chess} \cite{baxter}. Researchers managed to cast the objectives into summations of rewards, by assigning every reward a zero value except for the terminal reward. Problems seeking fast task completions form another class of examples, such as maze navigation or the mountain-car control task \cite{mountaincar}. Researchers cast the objectives as cumulative rewards by assigning a penalty for each action the agent takes before reaching the destination \cite{kaelbling}. There are also researches on objectives that are not easily cast into summations, such as the objectives as the \emph{average reward} \cite{blackwell,ferguson,rlearning,sridhar,tadepalli,proper,chenyu}. To optimize the average reward, besides computing the summation of rewards, the number of steps is either tracked explicitly \cite{ferguson}, or taken to the limit at infinity (for cyclic non-terminating MDPs) \cite{rlearning,sridhar}. Regardless, the summation operation in the Bellman optimality equation remains in these proposed algorithms. There have been two works \cite{maxreward1, maxreward2} exploring maximum-reward objectives, with applications on financial derivatives and medicine design. These works recognize the possibility of modifying the Bellman optimality equation, however their scopes are restricted to the maximum-reward objective formulation, instead of generalizing to a larger class of objective functions or proposing universal conditions for convergence. Furthermore, for MDPs whose state transition is a stochastic function of the input, the convergence to the global optimal policy cannot be guaranteed for the approach in \cite{maxreward1} and \cite{maxreward2}.

In this paper, we generalize the optimal control and reinforcement learning objectives to a variety of functions over the intermediate rewards. To optimize the generalized objectives, we exploit the flexibility in the Bellman optimality equation and modify it accordingly to the generalized objective functions. Specifically, we replace the summation operation in the Bellman optimality equation by new operations catering to the non-cumulative objective functions. Through this approach, we can readily adapt the existing optimal control or reinforcement learning algorithms to optimizing non-cumulative objectives, without needing to re-engineer a new set of artificial rewards just to cast the objectives into a summation of rewards. Furthermore, we provide the theoretical analysis on the generalized Bellman updates, and propose sufficient conditions on the form of the new operation as well as the assumptions on the MDP under which the global optimality of the converged value function and the corresponding greedy policy can be guaranteed. 

By expanding the possibilities of the objective functions, we are now able to solve problems with objectives that are intrinsically non-cumulative. For experiments, we focus on the \emph{bottleneck} objective: the objective as the minimum reward of all intermediate rewards. To optimize bottleneck objectives, we replace the summation operation in the Bellman optimality equation by the minimization operation, and apply the generalized Bellman update rule to learn the value function. In numerical simulations, we first re-formulate two classical reinforcement learning problems: the \emph{CartPole} problem \cite{cartpole} and the Atari game \emph{Breakout}, with bottleneck objectives. Through optimizing these problems with the proposed generalized Bellman updates, we obtain competitive performances by policies with different strategies from the classical solutions. 

We further experiment on two network communication applications with bottleneck objectives: the problem of finding the single-path maximum flow on a directed graph as an optimal control task, as well as joint routing and spectrum access over a wireless ad hoc network as a reinforcement learning problem. The proposed approach achieves excellent performances on both problems that are otherwise difficult to solve using the conventional formulation and learning algorithms. Specifically, for the wireless ad hoc network problem, a prior work \cite{drl_routing} has explored the Monte-Carlo estimation approach for learning the value function. In contrast, the proposed generalized update rule allows for the adaptation of the highly efficient temporal difference learning technique \cite{temporaldifference} to the generalized objective formulation, which results in noticeably faster and more stable learning progress. Furthermore, as the wireless ad hoc network problem is essentially a multi-agent reinforcement learning (MARL) problem, the results obtained also suggest that the proposed approach is readily compatible and effective under the multi-agent reinforcement learning setting.

The rest of the paper is organized as follows. In Section~\ref{sec:prob}, we introduce the general problem description on optimizing non-cumulative objectives, as well as several examples where non-cumulative objectives are applicable. In Section~\ref{sec:method}, we formally propose the method of the generalized Bellman update rules, and provide theoretical convergence and optimality analysis. We provide the detailed problem formulations on several example applications, and elaborate on how the proposed generalizations can be applied to optimizing such specific applications in Section~\ref{sec:specificmethod}, followed by the numerical simulations and analysis of the results in Section~\ref{sec:simulations}. Lastly, we draw conclusions in Section~\ref{sec:conclusion}.

\section{Generalized Optimal Control \& Reinforcement Learning Formulation}\label{sec:prob}
\subsection{Conventional Formulation}\label{sec:prob_I}
Let $\mathcal{S}$ and $\mathcal{A}$ denote the state space and the action space of an MDP. At time step $t$, the agent observes a state $s_t\in\mathcal{S}$, executes an action $a_t\in\mathcal{A}$, and receives a reward $r_t\in\mathcal{R}$ while transiting to the next state $s_{t+1}\in\mathcal{S}$. We use $\{p_{R_t|S_t,A_t}(r_t|s_t,a_t)\}_{t=1,2\dots}$ and $\{p_{S_{t+1}|S_t,A_t}(s_{t+1}|s_t, a_t)\}_{t=1,2\dots}$ to denote the reward distribution and the state transition distribution of the MDP, but often omit the subscripts for notational simplicity, e.g., as in $\{p(r_t|s_t,a_t)\}_{t=1,2\dots}$ and $\{p(s_{t+1}|s_t,a_t)\}_{t=1,2\dots}$. In most of the literature, the objective is defined as the summation of all intermediate rewards the agent received along the process: 
\begin{align}\label{equ:cumulative}
    u &= r_1+\gamma r_2+\gamma^2 r_3+\dots\:\:,
\end{align}
where $\gamma\in(0,1)$ is the \emph{discount factor} to encourage the agent focuses more on rewards closer in time. The study of control (when both the reward distribution $p(r_t|s_t,a_t)$ and the state transition distribution $p(s_{t+1}|s_t,a_t)$ are known) or reinforcement learning (when neither $p(r_t|s_t,a_t)$ nor $p(s_{t+1}|s_t,a_t)$ is known) is to find a policy $\pi$ for the agent to select actions based on states as $a_t\sim\pi(s_t),\forall t$, such that $u$ in \cref{equ:cumulative} is optimized. 

Corresponding to \cref{equ:cumulative}, the \emph{value function} is defined as the future cumulative rewards the agent expect to receive under a specific policy. Let $\mathcal{V}=\{(s,a)\:|\:s\in\mathcal{S}, a\in\mathcal{A}\}$ denote the set of all possible state-action pairs. The value function $Q^\pi\in\mathcal{R}^{|\mathcal{V}|}$ is a vector containing the future cumulative rewards expected starting from each $(s,a)$ tuple, with: 
\ifOneColumn
\begin{align}
    Q^\pi(s_t, a_t) 
=&\: \mathbb{E}_{\substack{\{p(r_{t'}|s_{t'},a_{t'})\}_{t'=t,t+1\dots} \\ \{p(s_{t'+1}|s_{t'},a_{t'})\}_{t'=t,t+1\dots}\\ \{a_{t'+1}\sim\pi(s_{t'+1})\}_{t'=t,t+1\dots} }}\left[r_t+\gamma r_{t+1}+\gamma^2 r_{t+2}+\dots | s_t,a_t \right] \label{equ:Q_original} \\
     =&\: \mathbb{E}_{\substack{p(r_t|s_t,a_t)\\p(s_{t+1}|s_t,a_t)\\a_{t+1}\sim\pi(s_{t+1})}}[r_t+\gamma Q^\pi(s_{t+1},a_{t+1}) | s_t,a_t]\:. \label{equ:bellman_expect}
\end{align}
\else
\begin{align}
    Q&^\pi(s_t, a_t) \nonumber \\
=&\mathbb{E}_{\substack{\{p(r_{t'}|s_{t'},a_{t'})\}_{t'=t,t+1\dots} \\ \{p(s_{t'+1}|s_{t'},a_{t'})\}_{t'=t,t+1\dots}\\ \{a_{t'+1}\sim\pi(s_{t'+1})\}_{t'=t,t+1\dots} }}\left[r_t+\gamma r_{t+1}+\gamma^2 r_{t+2}+\dots | s_t,a_t \right] \label{equ:Q_original} \\
     =&\mathbb{E}_{\substack{p(r_t|s_t,a_t)\\p(s_{t+1}|s_t,a_t)\\a_{t+1}\sim\pi(s_{t+1})}}[r_t+\gamma Q^\pi(s_{t+1},a_{t+1}) | s_t,a_t]\:. \label{equ:bellman_expect}
\end{align}
\fi
As shown in \cite{bertsekas}, for stationary single-agent fully-observable MDPs, there exists a deterministic global-optimal policy $\pi^*$, with its value function denoted as $Q^*$, with the following relationship:
\begin{align}\label{equ:greedy_policy}
    \pi^*(s_t)=\text{argmax}_aQ(s_t,a)
\end{align}
Essentially, $\pi^*(s_t)$ is a \emph{deterministic distribution} with all its probability density on the single action that maximizes $Q(s_t,a_t)$. Therefore, $\pi^*$ is commonly referred to as a \emph{greedy policy}.

For optimal control, $Q^*$ can be computed by \cref{equ:bellman_expect} with $\pi$ being the global optimal greedy policy $\pi^*$, leading to the \emph{Bellman optimality equation}: 
\ifOneColumn
\begin{align}\label{equ:bellman_optimality}
    Q^*(s_t, a_t) =\mathbb{E}_{\substack{p(r_t|s_t,a_t)\\p(s_{t+1}|s_t,a_t)}}[r_t+\gamma\max_{a_{t+1}}Q^*(s_{t+1},a_{t+1}) | s_t,a_t]\:. 
\end{align}
\else
\begin{align}\label{equ:bellman_optimality}
    Q&^*(s_t, a_t) \nonumber\\
    &=\mathbb{E}_{\substack{p(r_t|s_t,a_t)\\p(s_{t+1}|s_t,a_t)}}[r_t+\gamma\max_{a_{t+1}}Q^*(s_{t+1},a_{t+1}) | s_t,a_t]\:. 
\end{align}
\fi

Meanwhile, for reinforcement learning, $Q^*$ is learned through iterative updates of sample-based approximations to \cref{equ:bellman_optimality}, known as the \emph{Bellman update}:
\begin{align}\label{equ:bellman_update}
    Q(s_t, a_t) \leftarrow r_t+\gamma\max_{a_{t+1}}Q(s_{t+1}, a_{t+1})\:,
\end{align}
where the superscript on $Q$ is dropped, since during these updates, $Q$ does not necessarily correspond to the value function of any policy. We note that in Bellman updates, as shown by \cref{equ:bellman_update}, the updated estimations for $Q$ are obtained through bootstrapping from the current estimations. This learning technique is commonly known as \emph{temporal difference learning} \cite{temporaldifference}, which enjoys low estimation variance and high learning efficiency. \cref{equ:bellman_update} is used directly in the value-based algorithms such as SARSA \cite{sarsa}, Q-learning \cite{qlearning}, with the process commonly referred to as \emph{value iteration}; and policy-based algorithms such as the class of Actor-Critic methods \cite{actorcritic}. 

\subsection{Generalized Non-Cumulative Objectives}\label{sec:prob_II}
While it is proper to express the objective as \cref{equ:cumulative} in many scenarios, there exist applications where the objective $u$ is intrinsically some other function over the intermediate rewards. In this paper, to generalize the class of objectives that can be optimized, we formulate the objectives as \emph{general functions} over intermediate rewards: 
\begin{align}\label{equ:general_obj}
    u=f(r_1, r_2,r_3,\dots)\: .
\end{align}

Examples for such objectives can be seen from a wide variety of problems, which include, but are not limited to, the following classes of problems: 
\begin{itemize}
\item The \emph{bottleneck} of the intermediate rewards along the process, which fits into the large class of \emph{max-min optimization problems} \cite{drl_routing,haohang,manijeh,hailiang}. Among these max-min optimizations, the \emph{network routing} problems are perhaps the most standout examples.
\item The \emph{largest} reward among the intermediate rewards along the process \cite{maxreward1, maxreward2}. 
\item The \emph{harmonic mean} of the intermediate rewards along the process, such as the average traveling velocity, electrical resistance in circuits, density of mixture. It has also been used in wireless communications as a measure of fairness among users \cite{mina}.
\end{itemize}

Among various non-cumulative objectives, the objective of the \emph{bottleneck reward} is particularly prevalent. An important class of problems with bottleneck objectives are the network routing problems. Consider a data flow in a communication network consisting of multiple links, the highest rate the flow supports is the rate of the bottleneck link (i.e. the link with the lowest rate). Correspondingly, network routing problems are best formulated by the bottleneck objective. We describe such problems in detail in Section~\ref{sec:specificmethod}.

\section{Learning Algorithms with Generalized Bellman Updates}\label{sec:method}
This section aims to generalize optimal control and reinforcement learning to MDPs with non-cumulative objectives as \cref{equ:general_obj} by modifying the operation within the Bellman updates in \cref{equ:bellman_update}. We present sufficient conditions on the modified operation as well as assumptions on the underlying MDPs such that the Bellman updates still maintain the global optimal convergence property. Furthermore, we provide examples of frequently-encountered non-cumulative objectives with corresponding operations that satisfy the conditions for convergence. 

\subsection{Bellman Update with Generalized Operations}\label{sec:method_I}
Observing \cref{equ:bellman_update}, the update target of the new iteration consists of three fundamental elements:
\begin{enumerate}[label={(\alph*)}]
    \item Intermediate reward $r_t$, \label{enum:elem1}
    \item Value function at next state-action pair $Q(s_{t+1}, a_{t+1})$, \label{enum:elem2}
    \item Summation operation to combine \ref{enum:elem1} and \ref{enum:elem2}. \label{enum:elem3}
\end{enumerate}

In this paper, we explore substitutions of \ref{enum:elem3} in the Bellman optimality equation and its update rule by an alternative computational operation, which we refer to as the \emph{generalized Bellman update operation}, denote by $g(\cdot,\cdot)$. The operation takes \ref{enum:elem1} and \ref{enum:elem2} as the two arguments. As the result, we generalize \cref{equ:bellman_update} to the following form:
\begin{align}\label{equ:bellman_update_modified}
    Q(s_t, a_t) \leftarrow g\left(r_t, \gamma\max_{a_{t+1}}Q(s_{t+1}, a_{t+1})\right).
\end{align}

Through this generalized Bellman update operation, we are able to adapt the highly efficient temporal difference learning technique, as well as many popular reinforcement learning algorithms that based on it (e.g. SARSA, Q-learning, Actor-Critic), to optimizing the non-cumulative objectives, with minimal changes to these algorithms.

To determine which generalized objective functions $f(\cdots)$ as per \cref{equ:general_obj} can be optimized with such generalized Bellman updates, the first criterion is that the objective function needs to have \emph{optimal substructure} \cite{optimalsubstructure}, as a fundamental requirement of dynamic programming. Furthermore, for learning based algorithms with value function approximators (such as neural networks), it is desirable to have the value function $Q$ with fixed-dimension outputs from state to state (under most scenarios, the value function is a scalar function). This corresponds to the requirement that the objective function should be computable by iteratively updating a fixed number of statistics over its arguments (i.e. the intermediate rewards). When the two requirements are satisfied, we can deduce the proper operation $g(\cdot,\cdot)$ from the objective function $f(\cdots)$ on a case-by-case basis.

\subsection{Conditions for Convergence}\label{sec:method_II}
To facilitate the theoretical analysis, we denote each step of the value iteration by the function mapping $F^{\pi}:\mathcal{R}^{|\mathcal{V}|}\to\mathcal{R}^{|\mathcal{V}|}$. The superscript $\pi$ indicates that the policy $\pi$ is used for action selection in the one-step look ahead target computation. Correspondingly, we have:
\begin{align}\label{equ:bellman_update_modified_mapping}
    (F^\pi Q)(s_t, a_t) = \mathbb{E}_{\substack{p(r_t|s_t,a_t)\\ p(s_{t+1}|s_t,a_t)\\
    a_{t+1}\sim\pi(s_{t+1})}}g\Big(r_t, \gamma Q(s_{t+1}, a_{t+1})\Big),
\end{align}
where the expectation is understood as conditioned under $(s_t, a_t)$. When the deterministic greedy policy as derived from the current $Q$ is used, the value iteration is denoted by $F^*$ as follows:
\begin{align}\label{equ:bellman_update_modified_mapping1}
    (F^*Q)(s_t, a_t) = \mathbb{E}_{\substack{p(r_t|s_t,a_t)\\ p(s_{t+1}|s_t,a_t)}}g\left(r_t, \gamma\max_{a_{t+1}}Q(s_{t+1}, a_{t+1})\right).
\end{align}

The original Bellman updates in \cref{equ:bellman_update} enjoy convergence to the global optimal value function, as shown in \cite{jaakkola,szepesvari}. To generalize this convergence property for the generalized updates as \cref{equ:bellman_update_modified_mapping1}, we present a sufficient condition on $g(\cdot,\cdot)$ for ensuring convergence to a unique value function in the following theorem.

\begin{theorem}\label{thm:1}
On a single-agent fully observable MDP, the series of value functions obtained from iteratively applying the generalized Bellman update rule $Q\leftarrow F^*Q$ as in \cref{equ:bellman_update_modified_mapping1} is guaranteed to converge to a unique convergence point in $\mathcal{R}^{|\mathcal{V}|}$ from any arbitrary starting point, if $g(\cdot,\cdot):\mathcal{R}\times\mathcal{R}\to\mathcal{R}$ satisfies the following condition:
\begin{align}\label{condition_1}
    |g(a,b)-g(a,c)|\leq|b-c|\quad\forall a,b,c\in\mathcal{R}
\end{align} 
\end{theorem}
The mathematical proof of this theorem is presented in Appendix~\ref{sec:appendix_I}.

Note that in \cref{thm:1}, we do not claim that the greedy policy resulted from the converged value function is the global optimal policy. Besides an additional condition we need on the operation $g(\cdot,\cdot)$ (to be introduced in the next subsection), the main reason is that after generalizing the Bellman update operation to $g(\cdot,\cdot)$, the value function learned through the value iteration process is no longer guaranteed to be the true expectation of the objective value as defined in \cref{equ:general_obj}, when the state transition functions and reward functions are stochastic. We elaborate on this observation in the following subsection.

\subsection{Suboptimality with Stochastic Transitions and Rewards}\label{sec:method_III}
With the generalized operation $g(\cdot,\cdot)$, we express the objective in (\ref{equ:general_obj}) with $g(\cdot,\cdot)$ as:
\begin{align}\label{equ:utility_generalized}
  u=f(r_1, r_2, r_3, \dots) = g\Big(r_1, \gamma g\big(r_2, \gamma g(r_3, \dots)\big)\Big)\:.
\end{align}
We have shown a condition on $g(\cdot,\cdot)$ for convergence in \cref{thm:1} for obtaining $Q^*$. Nonetheless, we observe that $Q^*$ does not necessarily recover the true expectation of $u$ when stochastic state transitions and rewards are considered. To illustrate this, consider an episode starting from the state $s_1$. Under the greedy policy $\pi^*$ derived from $Q^*$, we take the expectation over $p(r_t|s_t,a_t)$ and $p(s_{t+1}|s_t,a_t)$ on \cref{equ:utility_generalized}, which leads to:
\begin{align}\label{equ:before_exp}
\mathbb{E}&_{\substack{\{a_t=\pi^*(s_t)\}_{t=1,2...}\\ \{p(r_t|s_t,a_t)\}_{t=1,2...}\\ \{p(s_{t+1}|s_t,a_t)\}_{t=1,2...}}}[u(r_1, r_2, r_3, \dots)] \nonumber \\ &=\mathbb{E}_{\substack{\{a_t=\pi^*(s_t)\}_{t=1,2...}\\ \{p(r_t|s_t,a_t)\}_{t=1,2...}\\ \{p(s_{t+1}|s_t,a_t)\}_{t=1,2...}}}\bigg[g\Big(r_1, \gamma g\big(r_2, \gamma g(r_3, \dots)\big)\Big)\bigg]\:.
\end{align}
Meanwhile, starting from $t{=}1$, with the converged $Q^*$ obtained from the generalized Bellman updates, we have:
\begin{align}
\mathbb{E}&_{a_1=\pi^*(s_1)}\big[Q^*(s_1, a_1)\big] \nonumber \\
&=\mathbb{E}_{\substack{a_1=\pi^*(s_1)\\p(r_1|s_1,a_1)\\ p(s_2|s_1,a_1)\\a_2=\pi^*(s_2)}}\bigg[g\Big(r_1, \gamma Q^*(s_2, a_2)\Big)\bigg] \\
&=\mathbb{E}_{\substack{a_1=\pi^*(s_1)\\p(r_1|s_1,a_1)\\ p(s_2|s_1,a_1)\\a_2=\pi^*(s_2)}}\bigg[g\Big(r_1, \gamma\mathbb{E}_{\substack{p(r_2|s_2,a_2)\\ p(s_3|s_2,a_2)\\a_3=\pi^*(s_3)}}\big[g\big(r_2, \gamma Q^*(s_3, a_3)\big)\big]\Big)\bigg].\label{equ:Q_exp}
\end{align} 
Comparing \cref{equ:before_exp} and \cref{equ:Q_exp}, for $\mathbb{E}_{a_1=\pi^*(s_1)}\big[Q^*(s_1, a_1)\big]$ to be equal to the expectation of $u$ under $\pi^*$, $p(r_t|s_t,a_t)$, and $p(s_{t+1}|s_t,a_t)$, we require $g(\cdot,\cdot)$ to be exchangeable with $\mathbb{E}_{\pi^*}[\cdot]$, $\mathbb{E}_{p(r_t|s_t,a_t)}[\cdot]$, and $\mathbb{E}_{p(s_{t+1}|s_t,a_t)}[\cdot]$. With $\pi^*$ being the deterministic greedy policy as in \cref{equ:greedy_policy}, the operation $\mathbb{E}_{\pi^*}[\cdot]$ can always be exchanged with $g(\cdot,\cdot)$. However, if $p(r_t|s_t,a_t)$ or $p(s_{t+1}|s_t,a_t)$ is stochastic, $\mathbb{E}_{p(r_t|s_t,a_t)}[\cdot]$ or $\mathbb{E}_{p(s_{t+1}|s_t,a_t)}[\cdot]$ is not necessarily exchangeable with $g(\cdot,\cdot)$. In this case, \cref{equ:before_exp} and \cref{equ:Q_exp} can potentially evaluate to different values, and therefore $\pi^*$ derived from $Q^*$ may be suboptimal. 

Under this observation, in order to obtain a global optimality guarantee on the greedy policy $\pi^*$, we constrain the scope to deterministic MDPs. Furthermore, we introduce an additional condition on the generalized operation $g(\cdot,\cdot)$ in order to establish global optimality, as formally stated in the following theorem: 
\begin{theorem}\label{thm:2}
Given a non-cumulative objective function $u$ and its corresponding generalized Bellman update operation \mbox{$g(\cdot,\cdot):\mathcal{R}{\times}\mathcal{R}\to\mathcal{R}$} satisfying the condition \cref{condition_1} from \cref{thm:1}, let $Q^*$ denote the convergence point of the value iteration (from iteratively applying the generalized Bellman update rule as in \cref{equ:bellman_update_modified_mapping1}). For an MDP with deterministic $p(r_t|s_t,a_t)$ and $p(s_{t+1}|s_t,a_t)$, the greedy policy $\pi^*$ derived from $Q^*$ is guaranteed to be the global optimal policy, if $g(\cdot,\cdot)$ satisfies the following additional condition:
\begin{align}\label{condition_2}
b\geq c \quad\text{implies}\quad g(a,b)\geq g(a,c)\quad\forall a,b,c\in\mathcal{R}
\end{align}
\end{theorem}

The mathematical proof of this theorem is provided in Appendix~\ref{sec:appendix_II}.

We note that the assumptions on MDPs in \cref{thm:2} are satisfied by a large class of optimal control and reinforcement learning problems: e.g., board games including Go and Chess, a subset of Atari games, the class of network routing problems (such as the problems to be studied in Section~\ref{sec:specificmethod}), and so on.  

To summarize, given any general MDP, we may generalize its objective function and apply the generalized Bellman update as in \cref{equ:bellman_update_modified_mapping1} to try to learn its value function. If the generalized update operation satisfies the condition as in \cref{condition_1} in \cref{thm:1}, the value iteration is guaranteed to converge to a unique convergence point. Furthermore, if the underlying MPD satisfies the assumptions in \cref{thm:2}, and the update operation satisfies the condition \cref{condition_2}, the convergence point is the optimal value function and the greedy policy $\pi^*$ derived from the value function is guaranteed to be the global optimal policy.

\subsection{Examples of Generalized Objectives and Bellman Update Operations}
We introduce several widely applicable \mbox{non-cumulative} objectives, and present the corresponding modified Bellman update operations. In \mbox{Appendix B}, we provide the proofs that these operations satisfy the conditions in \cref{thm:1} and \cref{thm:2}.

\subsubsection{Bottleneck Reward Objective}\label{sec:bottleneck}
The objective $u$ is the minimum (i.e. \emph{bottleneck}) intermediate reward in the process:
\begin{align}\label{equ:utility_min}
    u(r_1, r_2, r_3, \dots) = \min(r_1,r_2,r_3,\dots).
\end{align}
The corresponding modified Bellman update operation is:
\ifOneColumn
\begin{align}\label{equ:Q_min}
    g(r_t, \gamma\max_{a_{t+1}}Q(s_{t+1}, a_{t+1}))=\min(r_t, \gamma\max_{a_{t+1}}Q(s_{t+1}, a_{t+1})),
\end{align}
\else
\begin{multline}\label{equ:Q_min}
    g(r_t, \gamma\max_{a_{t+1}}Q(s_{t+1}, a_{t+1}))=\\ \min(r_t, \gamma\max_{a_{t+1}}Q(s_{t+1}, a_{t+1})),
\end{multline}
\fi
where the discount factor $\gamma$ is useful for encouraging the agent to postpone the occurrences of negative rewards that often correspond to undesired or failure outcomes.

The proof that the bottleneck update operation satisfies both conditions in \cref{thm:1} and \cref{thm:2} is presented in Appendix~\ref{sec:appendix_III}.

\subsubsection{Maximum Reward Objective}
The objective $u$ is the maximum intermediate rewards within the process:
\begin{align}\label{equ:utility_max}
    u(r_1, r_2, r_3, \dots) = \max(r_1,r_2,r_3,\dots).
\end{align}
The corresponding modified Bellman update operation is:
\ifOneColumn
\begin{align}\label{equ:Q_max}
    g(r_t, \gamma\max_{a_{t+1}}Q(s_{t+1}, a_{t+1}))=\max(r_t, \gamma\max_{a_{t+1}}Q(s_{t+1}, a_{t+1})).
\end{align}
\else
\begin{multline}\label{equ:Q_max}
    g(r_t, \gamma\max_{a_{t+1}}Q(s_{t+1}, a_{t+1}))=\\ \max(r_t, \gamma\max_{a_{t+1}}Q(s_{t+1}, a_{t+1})).
\end{multline}
\fi

The proof that the maximum update operation \cref{equ:Q_max} satisfies both conditions in \cref{thm:1} and \cref{thm:2} follows the same logic as the the proof for the bottleneck update operation shown in Appendix~\ref{sec:appendix_III}.

\subsubsection{Harmonic Mean Reward Objective}
Assuming \mbox{$r_t>0,\:\forall t$}, and the process is always terminated after a fixed number of steps, the objective $u$ is the harmonic mean of all intermediate rewards within the process:
\begin{align}\label{equ:utility_harmonic}
    U(r_1, r_2,\dots,r_T) = \frac{1}{\frac{1}{r_1}+\frac{1}{r_2}+\frac{1}{r_3}+\dots+\frac{1}{r_T}}\:,
\end{align}
where we omit the constant reward count. Examples of such applications with harmonic mean objectives include:
\begin{itemize}
  \item Optimize average traveling speed over a trip consisting of a fixed number of intervals. 
  \item Minimize resistance in a circuit with a fixed number of resistors in parallel connection.
  \item Optimize mixture density (e.g. alloys) with a fixed number of selections on equal-weight components. 
\end{itemize}  
Although technically maximizing \cref{equ:utility_harmonic} is equally valid as minimizing the summation of inverse of rewards, we present it as an example of a non-cumulative objective function with a modified Bellman update operation (as shown below) that satisfies the proposed convergence conditions.  

The corresponding modified Bellman update operation is:
\ifOneColumn
\begin{align}\label{equ:Q_harmonic}
    g(r_t, \gamma\max_{a_{t+1}}Q(s_{t+1},a_{t+1})) = \frac{1}{\frac{1}{r_t}+\frac{1}{\gamma\max_{a_{t+1}}Q(s_{t+1}, a_{t+1}))}}\:.
\end{align}
\else
\begin{multline}\label{equ:Q_harmonic}
    g(r_t, \gamma\max_{a_{t+1}}Q(s_{t+1},a_{t+1})) = \\
	\frac{1}{\frac{1}{r_t}+\frac{1}{\gamma\max_{a_{t+1}}Q(s_{t+1}, a_{t+1}))}}\:.
\end{multline}
\fi

The proof that the harmonic mean update operation satisfies both conditions in \cref{thm:1} and in \cref{thm:2} is presented in Appendix~\ref{sec:appendix_IV}.

\section{Applications of Generalized Reinforcement Learning}\label{sec:specificmethod}
\subsection{Classical Reinforcement Learning Problems with Bottleneck Objectives}\label{sec:classic_problems}
We first re-examine classical reinforcement learning problems, formulated with the bottleneck objectives as introduced in \cref{sec:bottleneck}. In many classical optimal control and reinforcement learning applications, the agent's success is largely based on its ability to \emph{avoid failure or defeat}. This is particularly the case when the MDPs lack significant intermediate milestones or checkpoints, such as the \emph{CartPole} problem and the Atari game \emph{Breakout}. Instead of regarding such tasks as collecting as many rewards as possible, the agent can interpret the tasks with the equally valid strategy of \emph{avoiding the worst outcome} (corresponding to the lowest reward) as much as possible. 

Conventionally, both tasks are formulated with the cumulative objective, each with an incremental rewarding scheme. In the CartPole task, a positive reward is assigned to the agent for every timestep it maintains the pole in the upright position; while in Atari, a positive reward is assigned each time the agent breaks a brick with the bouncing ball.

To formulate the task with the bottleneck objective for such classical tasks, we assign a negative reward to the agent when an undesired or failure event occurs after executing a certain action. For the other actions that do not directly lead to the failure events, we simply assign a zero intermediate reward. In the CartPole task, the agent aims to control the cart to vertically balance the pole. When the pole falls outside a pre-defined angle range, a negative reward is assigned to the agent. Similarly, for the Atari game Breakout, the agent controls the movement of a paddle to catch and reflect a bouncing ball upwards to destroy layers of bricks located above. Each time the agent fails to catch the falling ball with the paddle, it is assigned a negative reward. With the discount factor $\gamma$ applied on rewards over time steps, the later the negative rewards occur, the higher the bottleneck objective is. 

By optimizing the bottleneck objective, the agent is able to learn \emph{alternative strategies} to these classical problems: For CartPole, the strategy is to prevent the pole from falling for as long as possible. For Breakout, the strategy is to keep the ball in play for a maximized duration through controlling the paddle to constantly catch and reflect the ball, which translates to, although not always most efficiently, maximizing the bricks destroyed and thus achieving competitive game scores. 

\subsection{Single-Path Maximum-Flow Routing with Bottleneck Objective on a Graph}\label{sec:flow_graph}
\subsubsection{Problem Setup}\label{sec:flow_graph_prob}
Consider a communication network modeled as a directed graph $G{=}(\mathcal{N}, \mathcal{E})$, where the set of nodes $\mathcal{N}$ corresponds to the transmission nodes, and the set of edges $\mathcal{E}$ corresponds to the communication links between the nodes. 

A single-path data flow is routed through the network, from a fixed source node $n_s\in\mathcal{N}$ towards a fixed destination node $n_t\in\mathcal{N}$. Each directed edge $e^{n_i\to n_j}\in\mathcal{E}$ from $n_i\in\mathcal{N}$ to $n_j\in\mathcal{N}$ represents the transmission link from $n_i$ to $n_j$, and is assigned with a link rate capacity $r(e^{n_i\to n_j})=r_{n_i\to n_j}$. We set $r_{n_i\to n_j}=0$ when there is no link from $n_i$ to $n_j$ in the network. The optimal routing problem is that of finding an ordered sequence of relay nodes as transmission hops, to form the route such that the bottleneck rate is maximized:
\ifOneColumn
\begin{align}\label{equ:graphproblem}
	\underset{n_1,n_2,\dots,n_m} {\text{maximize}}\: \min\big(r_{n_s\to n_1}, r_{n_1\to n_2},\dots,r_{n_{m-1}\to n_m}, r_{n_m\to n_t}\big),
\end{align}
\else
\begin{multline}\label{equ:graphproblem}
	\underset{n_1,n_2,\dots,n_m} {\text{maximize}}\: \min\big(r_{n_s\to n_1}, r_{n_1\to n_2},\dots,\\r_{n_{m-1}\to n_m}, r_{n_m\to n_t}\big),
\end{multline}
\fi
where $\{n_i\}_{i\in\{1\dots m\}}$ denote the $m$ relay nodes (with the number $m$ adjustable) forming the route of the flow.

\subsubsection{Generalized Optimal Control Solution}\label{sec:flow_graph_method}
To find the single-path maximum flow within a given network represented by a directed graph as described above, we formulate the routing process as an MDP: the agent moves along the frontier node of the route, and makes sequential decisions on the selection of the node for each hop, until the destination node is reached. For the state space $\mathcal{S}$, each state is uniquely identified by the frontier node the agent resides on. Specifically, we use ${\sf s}^{n_i}\in\mathcal{S}$ to denote the state that the current frontier node of the partially established route is node $n_i$. For the action space $\mathcal{A}$, we use ${\sf a}^{n_i{\to}n_j}\in\mathcal{A}$ to denote the action to move from node $n_i$ to node $n_j$. Lastly, as specified in the problem setup, $r_{n_i{\to}n_j}$ corresponds to the reward for the action ${\sf a}^{n_i\to n_j}$, which is the link rate capacity of the link from $n_i$ to $n_j$.

To optimize the objective (\ref{equ:graphproblem}), for each state and action pair (${\sf s}^{n_i}$, ${\sf a}^{n_i\to n_j}$), the generalized update is as follows:
\ifOneColumn
\begin{align}\label{equ:flow_graph_update_rule}
    Q({\sf s}^{n_i}, {\sf a}^{n_i\to n_j}) \leftarrow	
    \min\Big(r_{n_i\to n_j}, \gamma\max_{n_k}Q({\sf s}^{n_j}, {\sf a}^{n_j\to n_k})\Big).
\end{align}
\else
\begin{multline}\label{equ:flow_graph_update_rule}
    Q({\sf s}^{n_i}, {\sf a}^{n_i\to n_j}) \leftarrow	\\
    \min\Big(r_{n_i\to n_j}, \gamma\max_{n_k}Q({\sf s}^{n_k}, {\sf a}^{n_j\to n_k})\Big).
\end{multline}
\fi

From the converged $Q^*$, we obtain the global optimal greedy policy $\pi^*$ (guaranteed by the results in \cref{sec:method_II} and \ref{sec:method_III}), following which produces the flow route supporting the global maximal flow rate.

\subsection{Wireless Ad hoc Network Routing and Spectrum Access with Bottleneck Objective}\label{sec:physical_layer}
\subsubsection{Problem Setup}\label{sec:physical_layer_prob}
Consider the physical-layer routing problem as discussed in \cite{drl_routing}. In a wireless ad hoc network with a set of transmission nodes $\mathcal{N}$, a set of data flows $\mathcal{K}$ is to be established, each consisted of multiple hops with their own pairs of source and destination nodes. A set of frequency bands $\mathcal{B}$ is available for transmission, each with a bandwidth of \mbox{$W$ Hz}. We focus on two optimization tasks for these data flows: routing and spectrum access. The task of routing is to select an ordered list of intermediary relay nodes from $\mathcal{N}$ to form the route for each data flow. The task of spectrum access is to select a frequency band from $\mathcal{B}$ for the transmission of each hop in the route of each flow. We represent the route for flow $k\in\mathcal{K}$ as an ordered list denoted by
$\mathbf{n}^{(k)}$: 
\begin{align}\label{equ:nodelist}
	\mathbf{n}^{(k)}=(n^{(k)}_0, n^{(k)}_1, n^{(k)}_2\dots n^{(k)}_m, n^{(k)}_{m+1})\:,
\end{align}
where $n^{(k)}_0$ and
$ n^{(k)}_{m+1}$ represent the fixed source and destination node for flow $k$, and $\{n^{(k)}_i\}_{i\in\{1\dots m\}}$ represent the $m$ relay nodes (with the number $m$ adjustable) forming the route of flow $k$. We represent the spectrum access solution for flow $k$ as an ordered list denoted by $\mathbf{b}^{(k)}$, containing the selected frequency band of each hop: 
\begin{align}\label{equ:bandlist}
    \mathbf{b}^{(k)}=(b^{(k)}_1, b^{(k)}_2, b^{(k)}_3\dots b^{(k)}_{m+1})\:,
\end{align}
where $b^{(k)}_i \in\mathcal{B}$ denotes the frequency band selected for the $i$-th hop in the route of flow $k$, with $i\in\{1\dots m+1\}$. As the global topology of the ad hoc network is not available as inputs, the agents need to learn to infer the network topology during the routing process.

Consider a link from node $n_i$ to node $n_j$ over frequency band $b$. Let 
$h_{(n_i\to n_j,b)}\in\mathcal{C}$ denote its channel coefficient. The maximum transmission rate of this link is based on
the \emph{signal to interference plus noise ratio} (SINR) as follows:
\begin{subequations}\label{equ:rate}
\begin{align} 
\text{SINR}_{(n_i\to n_j,b)} =&\: \frac{x_{n_i,b}|h_{(n_i\to n_j,b)}|^2p}{\sum_{\substack{n_l\neq n_i,n_j \\ n_l\in\mathcal{N}}}x_{n_l,b}|h_{(n_l\to n_j,b)}|^2p + \sigma^2}\:, \label{equ:SINR} \\
r_{(n_i\to n_j,b)} =&\: W\log\left(1+\text{SINR}_{(n_i\to n_j,b)}\right)\:. \label{equ:instantRate}
\end{align}
\end{subequations}
where $p$ and $\sigma^2$ denote the
transmit power of each node and the background noise power on each frequency band. The binary control variable $x_{n_i,b}$ indicates whether the node $n_i$ is transmitting on the band $b$ or idle. The objective for each flow $u^{(k)}$ is the transmission rate it supports, which is the bottleneck link rate:
\begin{align}\label{equ:bottleneckrate}
	u^{(k)}= r_{\min}^{(k)} = \min_{i=0,1,2,\dots,m}r_{\left(n^{(k)}_i\to n^{(k)}_{i+1}, b^{(k)}_{i+1}\right)}\:.
\end{align}

The global objective $u$ over all data flows is then defined as the average of the bottleneck rates over all data flows:
\begin{align}\label{equ:globalbottleneckrate}
    u = \frac{\sum_{k\in\mathcal{K}}u^{(k)}}{|\mathcal{K}|}
\end{align}

\subsubsection{Generalized Reinforcement Learning Solution}\label{sec:physical_layer_method}
For the physical layer routing and spectrum access problem, we assign one agent per data flow, with each agent moving along the frontier node of its flow and making hop-by-hop decisions. With multiple data flows to be jointly optimized, this problem is essentially a multi-agent reinforcement learning problem, with higher complexity than the maximum flow routing problem on a graph as in Section~\ref{sec:flow_graph}. By optimizing this problem with the bottleneck objective formulation and the generalized Bellman updates, we demonstrate that the proposed approach is competitive and highly effective in the setting of multi-agent reinforcement learning.

For better parameter efficiency, we only train one set of parameters shared among all agents. We assume the wireless network is only partially observable to each agent, meaning $Q^*$ is no longer guaranteed to be global optimal. Nonetheless, as shown in later simulations, the corresponding $\pi^*$ is still competitive. We adopt the MDP formulation as in \cite{drl_routing}. At each step, each agent gathers 4 pieces of information on frequency band $b$ for each of its $c$ closest neighboring nodes: the \mbox{agent-to-neighbor} distance; the \mbox{neighbor-to-destination} distance; the angle between \mbox{agent-to-neighbor} and \mbox{agent-to-destination} directions; and the signal interference on the neighbor on \mbox{band $b$}. With this information, the agent forms the state $s$ on band $b$ with $s\in\mathbb{R}^{4c},\forall s\in\mathcal{S}$. For the action space $\mathcal{A}$, the agent has $c+1$ actions on band $b$: one action for connecting with each of the $c$ nodes via $b$, and one action for \emph{reprobing} \mbox{on $b$} (if none of the $c$ nodes is suitable). We use ${\sf s}^{(n_i,b)}$ to denote the state that the frontier node of the partially established flow is node $n_i$ and that the transmission to the next hop uses the band $b$. We use ${\sf a}^{(n_i\to n_j,b)}$ to denote the agent's action to establish the link from node $n_i$ to node $n_j$ using band $b$, which is assigned the reward as the rate of this link $r_{(n_i\to n_j,b)}$. During training, these rewards as link rates are computed after the routes are formed. 

As the bottleneck rate is not expressible as summations, \cite{drl_routing} uses the Monte-Carlo method \cite{montecarlo} for estimating the value function. The key improvement we propose over \cite{drl_routing} is to utilize the modified Bellman update rule for training the agents in the off-policy fashion, providing higher data efficiency, faster convergence, and better performances. Using \cref{equ:Q_min}, the generalized updates for training each agent are:
\ifOneColumn
\begin{align}\label{equ:Q_physical}
    Q({\sf s}^{(n_i,b)}, {\sf a}^{(n_i\to n_j,b)}) \leftarrow \min\big(r_{(n_i\to n_j,b)}, \gamma\max_{n_k,b'}Q({\sf s}^{(n_j,b')}, {\sf a}^{(n_j\to n_k,b')})\big).
\end{align}
\else
\begin{multline}\label{equ:Q_physical}
    Q({\sf s}^{(n_i,b)}, {\sf a}^{(n_i\to n_j,b)}) \leftarrow\\ \min\big(r_{(n_i\to n_j,b)}, \gamma\max_{n_k,b'}Q({\sf s}^{(n_j,b')}, {\sf a}^{(n_j\to n_k,b')})\big).
\end{multline}
\fi

After predicting $Q$ values for all frequency bands, the agent selects the action with the single highest $Q$ value among all bands to establish the new link, which specifies not only the optimal node as the next hop, but also the optimal frequency band for transmission to that node. 

\section{Simulations}\label{sec:simulations}
We experiment on the optimal control and reinforcement learning problems and compare solutions from the conventional Bellman updates and from our proposed generalized Bellman updates. We use following terms to refer to each algorithm:
\begin{itemize}
    \item \emph{Q-Min}: Optimal control solution or RL policy based on the value function obtained from the generalized Bellman update rule as \cref{equ:bellman_update_modified} and \cref{equ:Q_min}.
    \item \emph{Q-Sum}: Optimal control solution or RL policy based on the value function obtained from the conventional Bellman update rule as \cref{equ:bellman_update}.  
\end{itemize}

\subsection{Classical Reinforcement Learning Problems}\label{sec:exp_I}
We use the double-DQN architecture \cite{doubledqn} to model the agents. During training, the \mbox{$\epsilon$-greedy} policy with decaying $\epsilon$ is used for collecting experiences, along with prioritized experience replay \cite{prioritized} for sampling training batches in each update step. 

\subsubsection{CartPole Task}
To solve the CartPole task with the Q-Min algorithm, when the pole falls outside of the pre-defined angle range ($\pm12^{\circ}$ from the up-right position), we assign a negative reward of -1 to the agent. To encourage the agent to postpone negative reward occurrence, we use a discount factor $\gamma=0.95$ in \cref{equ:Q_min}. For learning with the Q-Sum algorithm, we follow the conventional incremental rewarding scheme that has been long used in this task.

\begin{figure}[!t]
\centering
\includegraphics[width=\columnwidth]{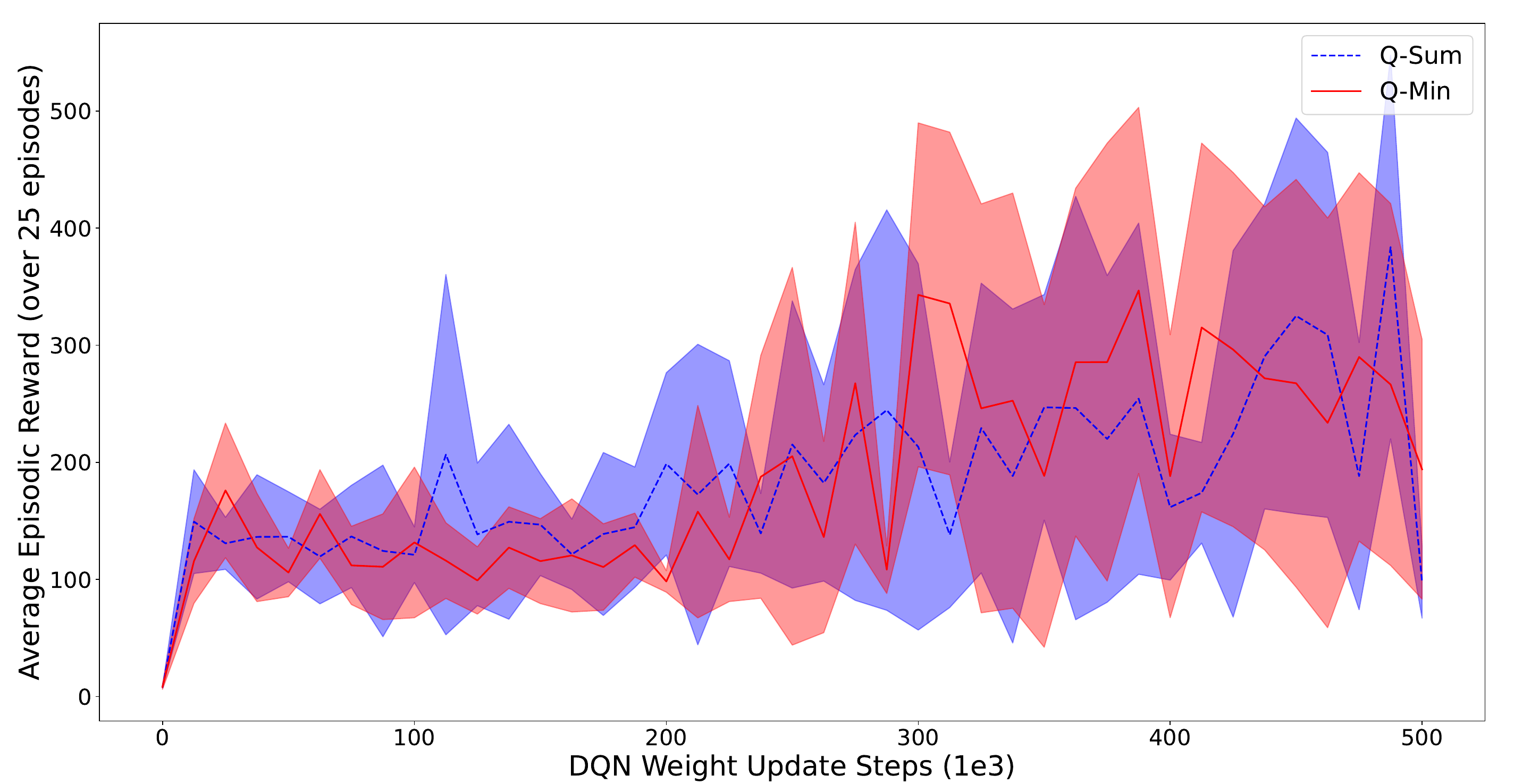}
\caption{Learning progress on CartPole: From 6 full training trails under 6 random seeds, we plot the mean as the line and the standard deviation as the shadow.}
\label{fig:cartpole}
\end{figure}

We illustrate the agent learning progress under both algorithms in \cref{fig:cartpole}, where we evaluate each agent's performance, averaged over 25 new episodes, after each 12500 update steps of training. We note that we stick with the conventional cumulative objective for CartPole as the performance metric when visualizing the learning progresses of both algorithms as shown in \cref{fig:cartpole}, which allows us to compare both algorithms directly. A competitive performance by the Q-Min algorithm on the cumulative objective would indicate that our alternative bottleneck objective is also a viable for formulating the task.   

As shown by the numerical results, besides the oscillations in both learning curves (as DQN is known for unstable learning), the Q-Min agent and the Q-Sum agent learn to balance the pole at a similar pace throughout training. The close results between the two algorithms validate that the bottleneck objective is indeed a suitable alternative to the CartPole objective formulation. 

\subsubsection{Atari Breakout Game}
To solve Atari with the proposed Q-Min algorithm, we utilize a simple reward scheme under Q-Min: we assign a negative reward of -1 to the agent each time it fails to catch the ball with the paddle, and set $\gamma=0.98$ in \cref{equ:Q_min} to encourage the agent to postpone such failure events. For learning with the Q-Sum algorithm, we follow the conventional incremental rewarding scheme originally built into the Atari game engine.

\begin{figure}[!t]
\centering
\includegraphics[width=\columnwidth]{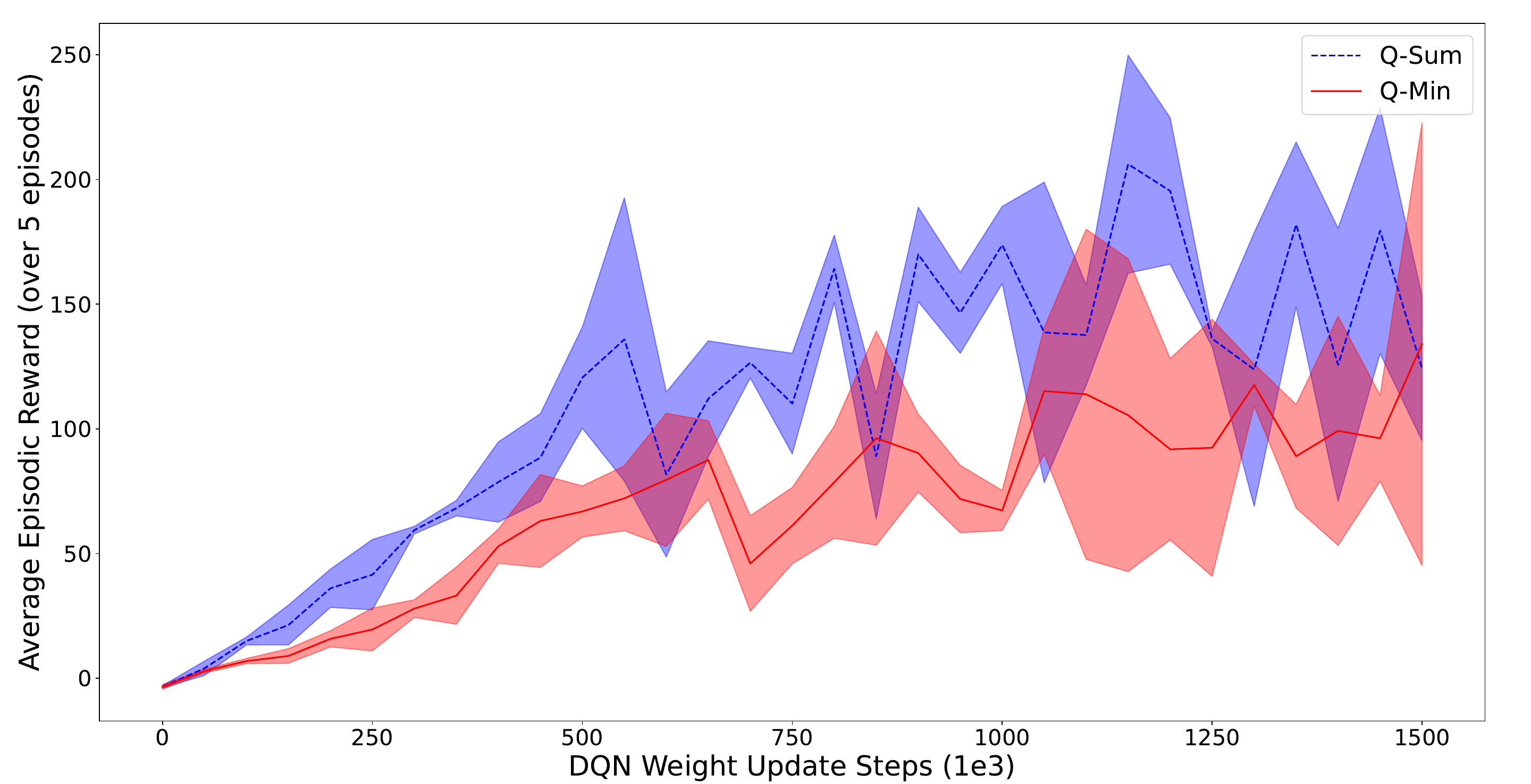}
\caption{Learning progress on Breakout: From 3 full training trails under 3 random seeds, we plot the mean as the line and the standard deviation as the shadow.}
\label{fig:breakout}
\end{figure}

We present the learning progress of \mbox{Q-Min} and \mbox{Q-Sum} in \cref{fig:breakout}, with each agent's performance evaluated and averaged over 5 new game runs, after each 50 thousand update steps of training. Similar to the learning progress visualization for CartPole, we also use the conventional cumulative objective for the original Breakout game as the performance metric when plotting the learning curves of both algorithms. 

Unlike in CartPole, the Q-Min agent shows a slightly slower learning progress and lower performance for Breakout. This is likely due to the Q-Min agent not learning the strictly optimal trajectories of redirecting the ball for hitting the most bricks, as its sole objective is to keep the ball in play. Nonetheless, even with simpler and more sparse rewards than the rewards used by Q-Sum, the Q-Min agent still manages to achieve relatively close performances to the Q-Sum agent, especially at the late training stage. The results illustrate the viability of interpreting Breakout with the bottleneck objective formulation.

We emphasize again that, specifically for these two classical problems, our goal is not to show that the proposed Q-Min algorithm is strictly superior to the conventional Q-Sum algorithm. After all, these two problems have long served as the canonical examples for the conventional reinforcement learning problems formulated with the cumulative objectives. Instead, we have shown that it is also valid to interpret and optimize these classical problems with the bottleneck objective formulation. Through learning with the proposed generalized Bellman update rule, the agent is capable of achieving the performances comparable with the results from the convention reinforcement learning approach as presented above. Essentially, When optimizing the agent under the bottleneck objective formulation, the agent learns an alternative game playing strategy for both CartPole and Breakout: to avoid or delay the failure event for as long as possible.

\subsection{Single-Path Maximum-Flow Routing on Graph}\label{sec:sim_flow_graph}
We consider the directed graph network shown in \cref{fig:graph_network}, and perform the Q-Min algorithm with \cref{equ:flow_graph_update_rule} until convergence. With the MDP in this problem being finite, we set $\gamma=1$, which simplifies the numerical results with $Q$ values precisely equal to the future bottleneck rates.

\ifOneColumn
\begin{figure}[!t]
\centering
\includegraphics[width=0.5\columnwidth]{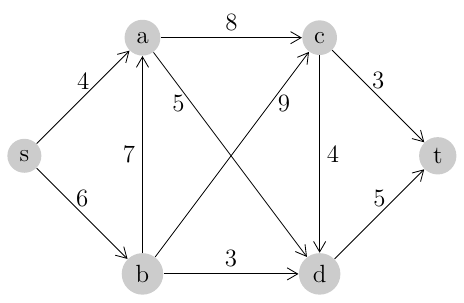}
\caption{The maximum network flow problem as modeled by a directed graph.}
\label{fig:graph_network}
\end{figure}
\else
\begin{figure}[ht]
\centering
\includegraphics[width=0.7\columnwidth]{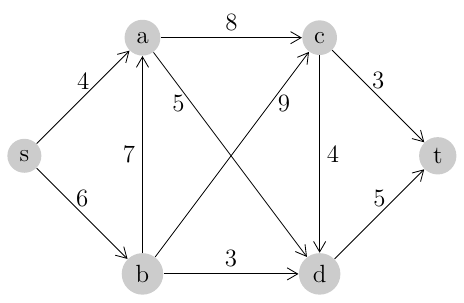}
\caption{The maximum network flow problem as modeled by a directed graph.}
\label{fig:graph_network}
\end{figure}
\fi

In Table~\ref{tab:result_flow_graph}, we present the iterations of both the Q-Min algorithm and the Q-Sum algorithm. In the first row of the table, we adopt the simplified notations for $Q$ values: we use $Q_{n_i\to n_j}$ to uniquely denote the state-action value function $Q(s_{n_i}, a_{(n_i,n_j)})$ in \cref{equ:flow_graph_update_rule}. We adopt synchronized iterations, where in each new iteration, the $Q$ value in the right-hand-side of \cref{equ:flow_graph_update_rule} comes from the previous iteration. All the iterations of value function updates are shown until convergence. 

For the Q-Min algorithm, it takes 4 iterations of the generalized Bellman updates to converge. From the resulted $Q^*$, we deduce the optimal policy $\pi^*$ producing the following optimal flow route:
\begin{align}
    s\to b\to a\to d\to t \: .
\end{align}
This route obtained supports a flow rate of 5, which is indeed the global optimal flow rate.

On the other hand, for the Q-Sum algorithm, the convergence speed is lower than Q-Min, as it takes 5 iterations of the regular Bellman updates to converge. Furthermore, the deduced optimal policy results in the following flow route:
\begin{align}
    s\to b\to a\to c\to d\to t \: .
\end{align}
This route supports a flow rate of 4, which is sub-optimal and inferior to the route obtained by the Q-Min algorithm.
\ifOneColumn
\begin{table*}[t!]
\centering
\caption{Q-Min Value Iterations on Graph Network Routing (shaded entries depict the routing selection under $\pi^*$).}
\begin{tabular}{c|c||cccccccccc}
\hline
\TBstrut Algorithm & Update Step & $Q_{d\to t}$ & $Q_{c\to t}$ & $Q_{c\to d}$ & $Q_{a\to c}$ & $Q_{a\to d}$ & $Q_{b\to d}$ & $Q_{b\to c}$ & $Q_{b\to a}$ & $Q_{s\to a}$ & $Q_{s\to b}$ \\ 
\hline
\multirow{5}*{Q-Min} & 0 & 0 & 0 & 0 & 0 & 0 & 0 & 0 & 0 & 0 & 0 \\
& 1 & 5 & 3 & 0 & 0 & 0 & 0 & 0 & 0 & 0 & 0 \\
& 2 & 5 & 3 & 4 & 3 & 5 & 3 & 3 & 0 & 0 & 0 \\
& 3 & 5 & 3 & 4 & 4 & 5 & 3 & 4 & 5 & 4 & 3 \\
& 4 (Converged) & \cellcolor{LightCyan}\textbf{5} & 3 & 4 & 4 & \cellcolor{LightCyan}\textbf{5} & 3 & 4 & \cellcolor{LightCyan}\textbf{5} & 4 & \cellcolor{LightCyan}\textbf{5} \\ 
\hline
\multirow{6}*{Q-Sum} & 0 & 0 & 0 & 0 & 0 & 0 & 0 & 0 & 0 & 0 & 0 
\\
& 1 & 5 & 3 & 4 & 8 & 5 & 3 & 9 & 7 & 4 & 6 \\
& 2 & 5 & 3 & 9 & 12 & 10 & 8 & 13 & 15 & 12 & 13 \\
& 3 & 5 & 3 & 9 & 17 & 10 & 8 & 18 & 19 & 16 & 21 \\
& 4 & 5 & 3 & 9 & 17 & 10 & 8 & 18 & 24 & 21 & 25 \\
& 5 (Converged) & \cellcolor{LightCyan}\textbf{5} & 3 & \cellcolor{LightCyan}\textbf{9} & \cellcolor{LightCyan}\textbf{17} & 10 & 8 & 18 & \cellcolor{LightCyan}\textbf{24} & 21 & \cellcolor{LightCyan}\textbf{30} \\
\hline
\end{tabular}
\label{tab:result_flow_graph}
\end{table*}
\else
\begin{table*}[hbt!]
\centering
\caption{Q-Min Value Iterations on Graph Network Routing (shaded entries depict the routing selection under $\pi^*$).}
\begin{tabular}{c|c||cccccccccc}
\hline
\TBstrut Algorithm & Update Step & $Q_{d\to t}$ & $Q_{c\to t}$ & $Q_{c\to d}$ & $Q_{a\to c}$ & $Q_{a\to d}$ & $Q_{b\to d}$ & $Q_{b\to c}$ & $Q_{b\to a}$ & $Q_{s\to a}$ & $Q_{s\to b}$ \\ 
\hline
\multirow{5}*{Q-Min} & 0 & 0 & 0 & 0 & 0 & 0 & 0 & 0 & 0 & 0 & 0 \\
& 1 & 5 & 3 & 0 & 0 & 0 & 0 & 0 & 0 & 0 & 0 \\
& 2 & 5 & 3 & 4 & 3 & 5 & 3 & 3 & 0 & 0 & 0 \\
& 3 & 5 & 3 & 4 & 4 & 5 & 3 & 4 & 5 & 4 & 3 \\
& 4 (Converged) & \cellcolor{LightCyan}\textbf{5} & 3 & 4 & 4 & \cellcolor{LightCyan}\textbf{5} & 3 & 4 & \cellcolor{LightCyan}\textbf{5} & 4 & \cellcolor{LightCyan}\textbf{5} \\ 
\hline
\multirow{6}*{Q-Sum} & 0 & 0 & 0 & 0 & 0 & 0 & 0 & 0 & 0 & 0 & 0 
\\
& 1 & 5 & 3 & 4 & 8 & 5 & 3 & 9 & 7 & 4 & 6 \\
& 2 & 5 & 3 & 9 & 12 & 10 & 8 & 13 & 15 & 12 & 13 \\
& 3 & 5 & 3 & 9 & 17 & 10 & 8 & 18 & 19 & 16 & 21 \\
& 4 & 5 & 3 & 9 & 17 & 10 & 8 & 18 & 24 & 21 & 25 \\
& 5 (Converged) & \cellcolor{LightCyan}\textbf{5} & 3 & \cellcolor{LightCyan}\textbf{9} & \cellcolor{LightCyan}\textbf{17} & 10 & 8 & 18 & \cellcolor{LightCyan}\textbf{24} & 21 & \cellcolor{LightCyan}\textbf{30} \\
\hline
\end{tabular}
\label{tab:result_flow_graph}
\end{table*}
\fi

\subsection{Wireless Ad hoc Network Routing and Spectrum Access}
\subsubsection{Experiment Settings}\label{sec:origin_setting}
We simulate wireless ad hoc networks in a 1000m${\times}$1000m region with $|\mathcal{K}|{=}3$ data flows and $|\mathcal{B}|{=}8$ frequency bands. We adopt the same specifications as in \cite{drl_routing} as we aim to compare results and illustrate the effectiveness of the proposed Q-Min algorithm. Specifically, we consider the short-range outdoor model ITU-1411 with a distance-dependent path-loss to model all wireless channels, over all frequency bands at 2.4GHz carrier frequency. Shadowing and fast-fading are not considered in the simulation setting. This corresponds to an outdoor environment (e.g., a rural or remote area), where the strengths of the wireless links are mostly functions of the distances between the transmitters and the receivers. We assume each of the $|\mathcal{B}|{=}8$ frequency bands has a 5MHz bandwidth for signal transmission. All antennas equipped at the transmission nodes have a height of 1.5m and 2.5dBi antenna gain. We assume a transmit power of 30dBm for all nodes and background noise at -130dBm/Hz. 

To generate realistic wireless network layouts, the node locations are randomly generated with varying node densities over the region. Specifically, we divide the 1000m${\times}$1000m network region into nine equal sub-regions, and randomly locate $(6, 8, 7, 6, 5, 10, 8, 9, 6)$ nodes within each of the nine sub-regions correspondingly. 

\subsubsection{Training Convergence Speed Comparison}\label{sec:sim_convergence}
We train each set of $|\mathcal{K}|{=}3$ agents with three algorithms: Q-Min, Q-Sum, and the algorithm by \cite{drl_routing}:
\begin{itemize}
    \item \emph{Q-MC}: RL policy based on the value function obtained from the Monte-Carlo episodic estimations of future bottleneck rewards, computed at the end of episodes.
\end{itemize}

We generate 380,000 wireless ad hoc network layouts for training the agents under each algorithm, under the following training schedule: 
\begin{itemize}
\item Initial 30,000 layouts are used for random routing on collecting initial experience.
\item The middle 300,000 layouts are used for the $\epsilon$-greedy policy based routing, with the $\epsilon$ value follows linear annealing from 1.0 to 0.1 throughout the training over these layouts. 
\item The final 50,000 layouts are used with $\epsilon=0$ for the final convergence stage.
\end{itemize}

We use the Dueling-DQN architecture \cite{dueling} to model all the agents, with the neural network specifications listed in \cref{tab:dueldqn}, same as in \cite{drl_routing}. Since the rewards as link rates are dense throughout the MDP, uniform sampling is sufficient for experience replay. We use $c{=}10$ as the number of neighbors the agent explores each time. The state inputs to the DQNs are therefore 40-component vectors, i.e. $s\in\mathcal{R}^{40},\forall s\in\mathcal{S}$.  

\begin{table}[t]
\centering
\caption{Dueling-DQN Neural Network Specifications}
\begin{tabular}{|c|c|c|}
\hline
\TBstrut Model Module & Hidden Layer & Neurons  \\
\hline
\multirow{2}*{\shortstack[c]{Feature Learning \\ Module}}
& \TBstrut 1st & 150 \\ \cline{2-3}
& \TBstrut 2nd & 150 \\ 
\hline
\multirow{2}*{\shortstack[c]{State-Value \\ Estimation Stream}} 
& \TBstrut 1st & 100 \\ \cline{2-3}
& \TBstrut 2nd & 1 (1 state value) \\ 
\hline
\multirow{2}*{\shortstack[c]{Action-Advantage \\ Estimation Stream}} 
& \TBstrut 1st & 100 \\ \cline{2-3}
& \TBstrut 2nd & 11 (11 actions) \\ 
\hline
\end{tabular}
\label{tab:dueldqn}
\end{table}

\begin{figure*}[!t]
\centering
\includegraphics[width=0.97\textwidth]{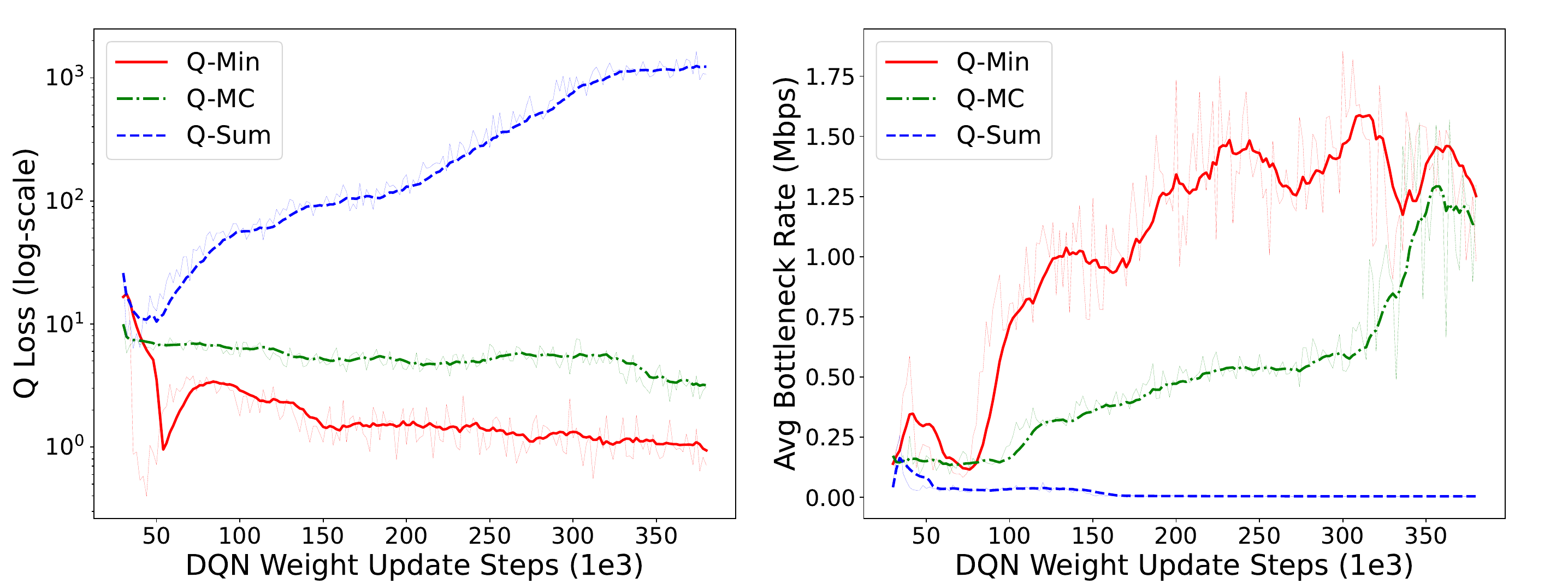}
\caption{Learning progress of various algorithms on the physical layer routing problem: The dotted lines are the performance values of all update steps; the solid lines are the smoothed curves, with each point being the average over the last 10 update steps.}
\label{fig:adhoc_training_curves}
\end{figure*}

For the same reasons as in \cref{sec:sim_flow_graph}, we set $\gamma=1$ in \cref{equ:Q_physical}. During training, we track both the mean-squared-errors for predictions on $Q$, and the routing performances over 100 newly generated network layouts at each update step of training. The training curves for all three algorithms are displayed in \cref{fig:adhoc_training_curves}.

Shown by the learning curves, the conventional Q-Sum agents collectively achieve the worst learning progress and simply fail to convergence on the $Q$ value estimations. While both the Q-Min agents and the Q-MC agents converge to comparable performances, the Q-Min agents enjoy a much faster convergence speed. This illustrates the advantage of the temporal difference learning over the Monte-Carlo method, which is made possible for non-cumulative objectives with the proposed generalized update rules. 

To better understand why Q-Min achieves noticeably faster convergence than Q-MC, we emphasize that the Monte-Carlo estimations used by Q-MC are highly affected by the random explorations especially at the early stage of training. Certain random explorations might lead to an extremely low bottleneck rate for the newly established route. This bottleneck rate is then used as the Monte-Carlo estimation on the value function during training the Q-MC agent. Thus, the value estimations learned by the Q-MC agents suffer from low qualities significantly at the beginning of training. 

On the other hand, with the proposed generalized update operation, the bottleneck objective can be estimated by the temporal difference learning technique as in the Q-Min algorithm. As an \emph{off-policy}\footnote[1]{An off-policy algorithm separates the policy that the value estimation is based on from the sampling policy, which is desired when the sampling policy is highly noisy (e.g. with many random explorations).} learning algorithm, the temporal difference learning estimations are much more resilient to the random explorations, since the estimation target is obtained through one-step bootstrapping on the already learned value function. Therefore, it is not a surprise to see the significant improvements on the training efficiency and convergence speed by the Q-Min agents.

\subsubsection{Performances on Bottleneck Rates}\label{sec:bottleneckrate}
We present in \cref{tab:sumrate} test results on the number of links established on each flow, as well as the achieved bottleneck flow rates for these data flows, over 1000 newly generated testing wireless ad hoc networks. Furthermore, for each method, we collect the bottleneck rates of all data flows over all testing wireless networks, and present the cumulative distribution function (CDF) of these bottleneck rates in \cref{fig:flowrates}. As shown by both the statistics and the distributions of the flow rates, the Q-Min agents achieve the best routing results, whereas the Q-Sum agents perform the worst by a large margin, while having much higher numbers of links over the established data flows. 

\begin{table}[t]
\centering
\caption{Average Bottleneck Flow Rate Performances}
\begin{tabular}{ccc}
\hline
\TBstrut Methods & Average \# Links per Flow & Average Bottleneck Rate \\
\hline
\TBstrut Q-Min & 13.0 & 1.76 Mbps \\
\TBstrut Q-MC & 13.8 & 1.42 Mbps \\
\TBstrut Q-Sum & 31.2 & 0.21 Mbps \\
\hline
\end{tabular}
\label{tab:sumrate}
\end{table}

\ifOneColumn
\begin{figure*}[!t]
\centering
\includegraphics[width=0.9\columnwidth]{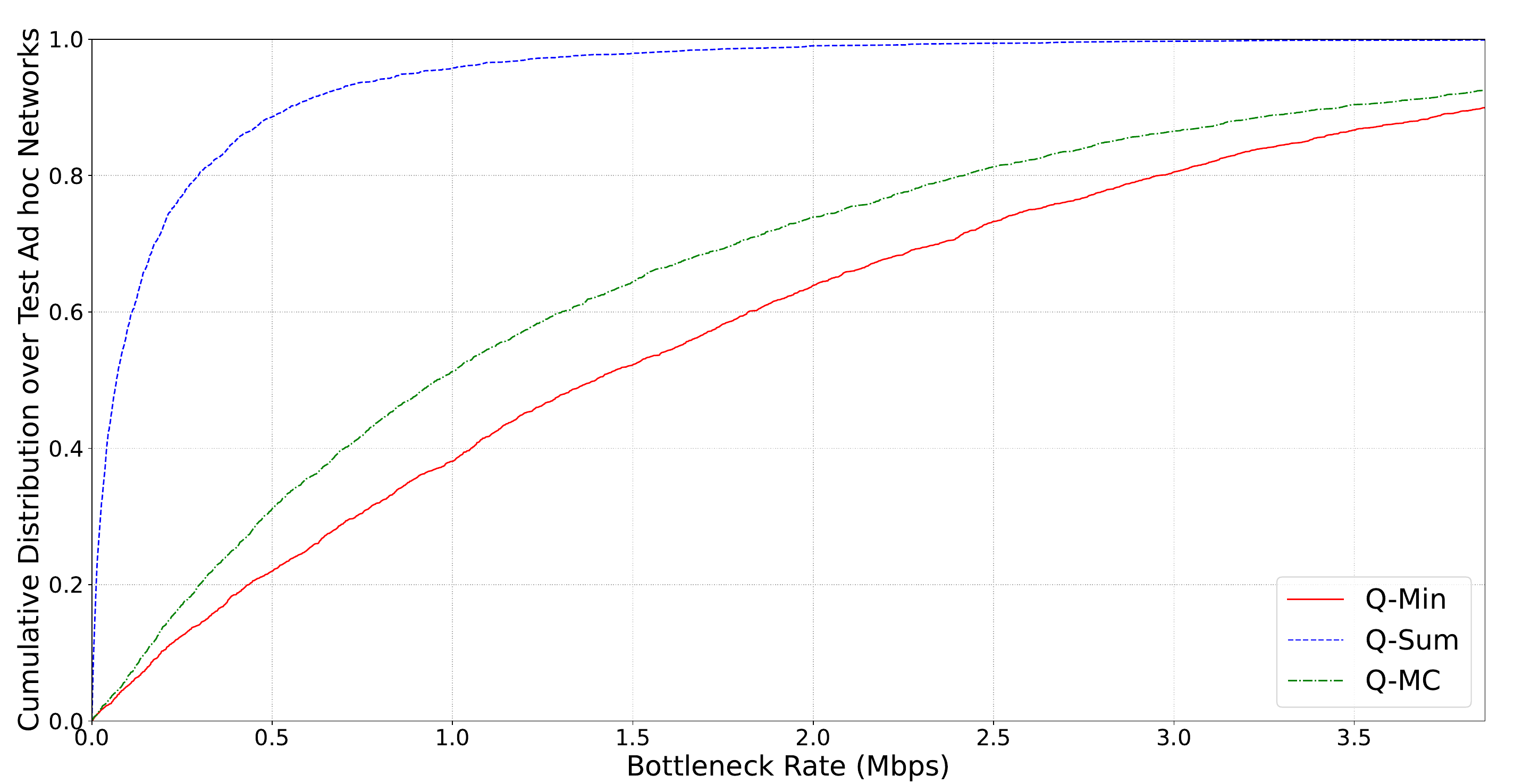}
\caption{Cumulative distribution function of flow bottleneck rates in 1000 testing wireless ad hoc networks.}
\label{fig:flowrates}
\end{figure*}
\else
\begin{figure*}[ht]
\centering
\includegraphics[width=1.13\columnwidth]{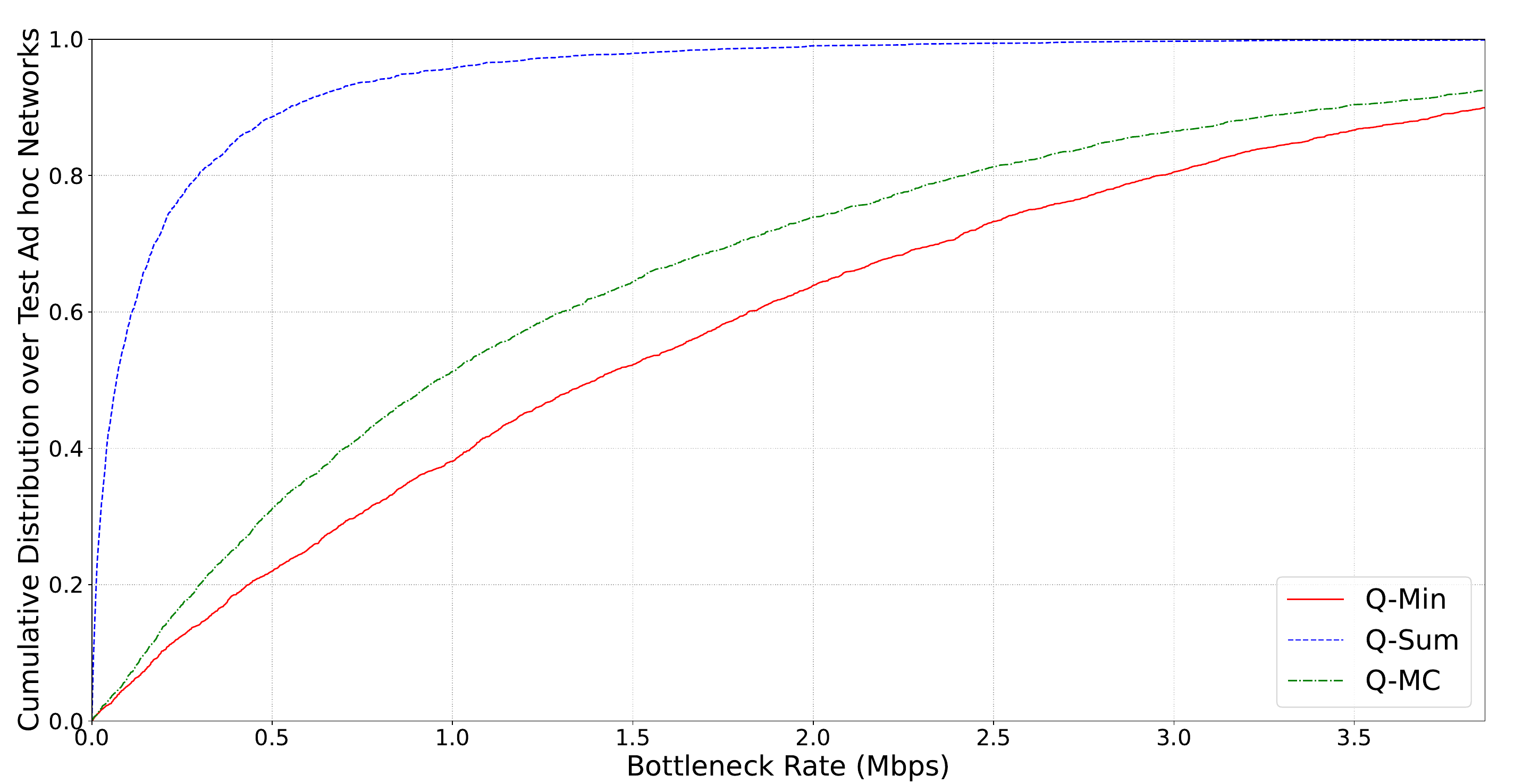}
\caption{Cumulative distribution function of flow bottleneck rates in 1000 testing wireless ad hoc networks.}
\label{fig:flowrates}
\end{figure*}
\fi

We visualize the optimized routes by each RL algorithm over a random wireless ad hoc network in \cref{fig:routes}. The Q-Min agents learn to establish links with medium lengths. This policy ensures a certain level of channel strength for the bottleneck links, without constructing too many links to avoid excessive interference which is detrimental to the bottleneck link rates. Furthermore, the Q-Min agents also learn to spatially spread out data flows as well as the frequency bands used among links for effective interference mitigation. 

On the other hand, the Q-Sum agents learn the policy that connects unnecessarily many short links to form routes, neglecting the importance of the bottleneck link within each flow. Evidently, the conventional reinforcement learning formulation is unsuitable for solving such routing problems. For this reason, in application fields such as network communications, generalizing the objective function and its learning rule through our proposed approach is an essential optimization technique. 

\ifOneColumn
\begin{figure*}[!t]
\centering
\includegraphics[width=\textwidth]{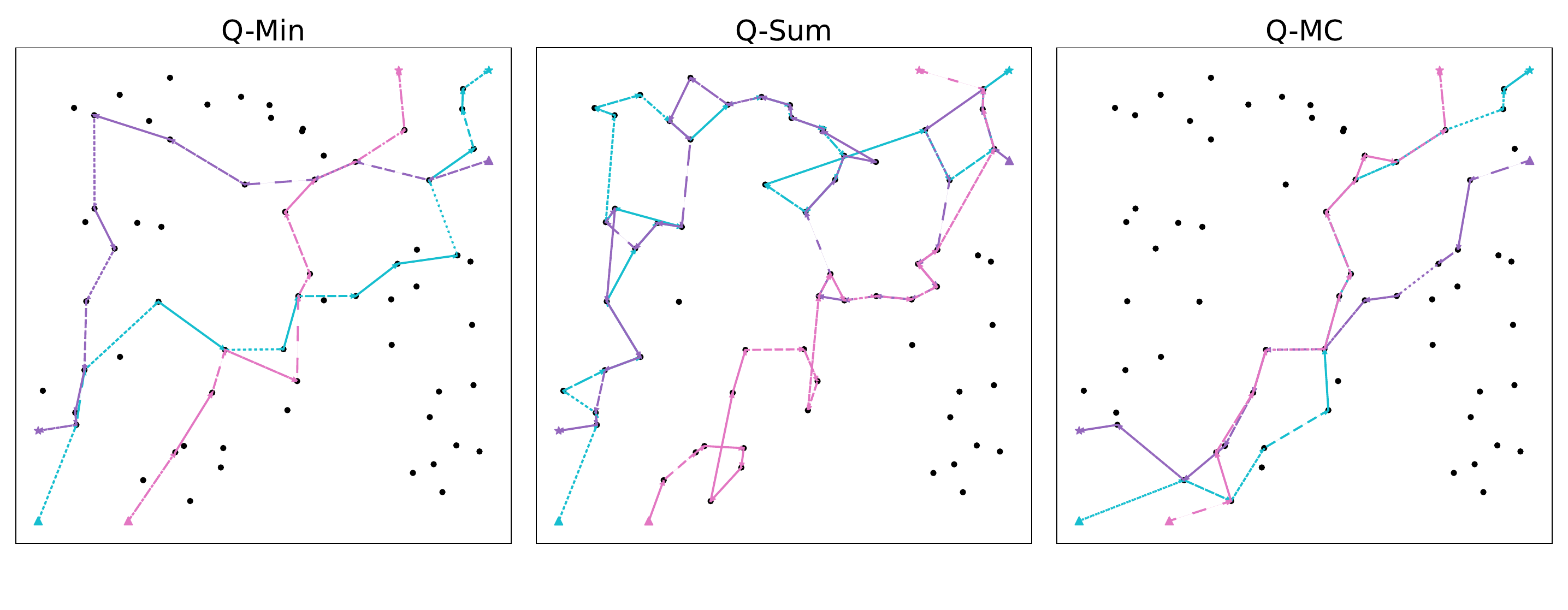}
\caption{Routes for data flows (data flows differentiated by colors; frequency bands differentiated by linestyles).}
\label{fig:routes}
\end{figure*}
\else
\begin{figure*}[ht]
\centering
\includegraphics[width=0.93\textwidth]{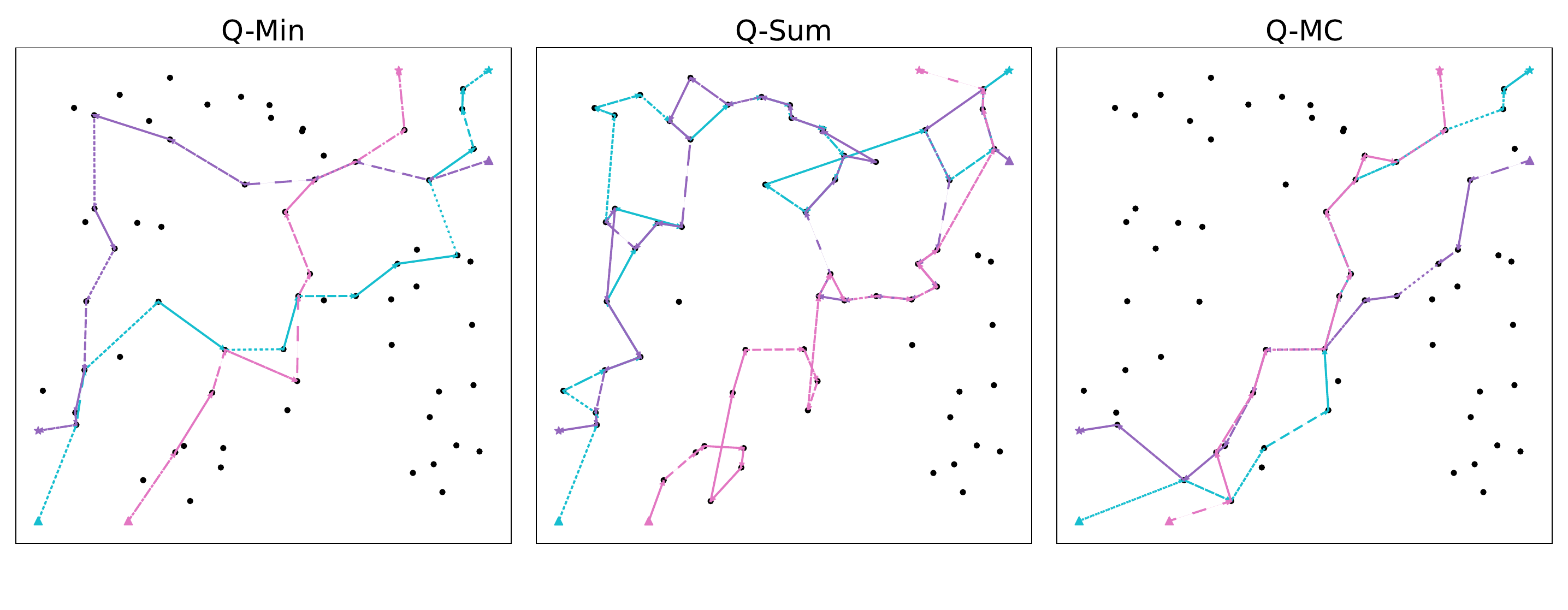}
\caption{Routes for data flows (data flows differentiated by colors; frequency bands differentiated by linestyles).}
\label{fig:routes}
\end{figure*}
\fi

\section{Conclusion}\label{sec:conclusion}
This paper recognizes the possibilities of formulating optimal control or reinforcement learning objectives as non-cumulative functions over rewards, and generalizes existing algorithms to optimizing such objectives. Specifically, we explore the generalized operations in the Bellman update rule, for which we provide the global convergence conditions with mathematical proofs. We also recognize the assumptions required on the MDP state transitions and reward functions for ensuring the global optimality on the obtained policies. With the generalized objectives and learning algorithms, we are able to unveil alternative strategies to classical optimal control or reinforcement learning problems, and more importantly, realize the possibilities for solving new problems with intrinsically non-cumulative objectives, which are frequently encountered in the fields such as network communications. This opens up directions for a broader range of applications for optimal control and reinforcement learning techniques.

 \appendices

\section{Proof of \cref{thm:1}}\label{sec:appendix_I}
\begin{lemma}\label{lemma:1}
If $g(\cdot,\cdot)$ satisfies the condition \cref{condition_1} in \cref{thm:1}, then the generalized value function update $F^*$ as in \cref{equ:bellman_update_modified_mapping1} is a contraction mapping.
\end{lemma}
\begin{proof}
In the following mathematical expressions, for the simplicity of notations, we use $p(r_t,s_{t+1}|s_t,a_t)$ as the shorthand notation for the joint distribution of $p_{R_t|S_t,A_t}(r_t|s_t,a_t)$ and $p_{S_{t+1}|S_t,A_t}(s_{t+1}|s_t,a_t)$. We also assume $p(r_t,s_{t+1}|s_t,a_t)$ is a discrete distribution. For continuous distribution, the proof still holds with summations substituted by integrations when computing the expectations. 

For any pair of value functions $\forall\: Q^1, Q^2 \in\mathcal{R}^{|\mathcal{V}|}$, we have:
\allowdisplaybreaks
\ifOneColumn
\begin{align}
    \lVert F&^*Q^1-F^*Q^2\rVert_{\infty} \nonumber\\
    =& \max_{s_t,a_t}\left| (F^*Q^1)(s_t,a_t)-(F^*Q^2)(s_t,a_t)\right| \\
    =& \max_{s_t,a_t}\Big|\sum_{r_t,s_{t+1}}p(r_t,s_{t+1}|s_t,a_t)g(r_t, \gamma \max_{a_{t+1}}Q^1(s_{t+1}, a_{t+1})) \nonumber \\
    &\quad\quad\quad\quad\quad\quad\quad\quad\quad\quad\quad\quad -\sum_{r_t,s_{t+1}}p(r_t,s_{t+1}|s_t,a_t)g(r_t, \gamma\max_{a_{t+1}}Q^2(s_{t+1}, a_{t+1}))\Big|  \\
    =& \max_{s_t,a_t} \Big|\sum_{r_t,s_{t+1}}p(r_t,s_{t+1}|s_t,a_t)\big[g(r_t, \gamma \max_{a_{t+1}}Q^1(s_{t+1}, a_{t+1}))-g(r_t, \gamma \max_{a_{t+1}}Q^2(s_{t+1}, a_{t+1}))\big]\Big| \\
    \leq& \max_{s_t,a_t}\sum_{r_t,s_{t+1}}p(r_t,s_{t+1}|s_t,a_t)\Big| g(r_t, \gamma \max_{a_{t+1}}Q^1(s_{t+1}, a_{t+1})) -g(r_t, \gamma \max_{a_{t+1}}Q^2(s_{t+1}, a_{t+1}))\Big|\\
    \leq& \: \max_{s_t,a_t}\sum_{r_t,s_{t+1}}p(r_t,s_{t+1}|s_t,a_t)\big|\gamma\max_{a_{t+1}}Q^1(s_{t+1}, a_{t+1})  -\gamma\max_{a_{t+1}}Q^2(s_{t+1}, a_{t+1})\big| \label{equ:use_lemma_1}\\
    \leq& \: \gamma\max_{s_t,a_t}\sum_{r_t,s_{t+1}}p(r_t,s_{t+1}|s_t,a_t)\max_{a_{t+1}}\big| Q^1(s_{t+1}, a_{t+1}) -Q^2(s_{t+1}, a_{t+1})\big| \label{equ:supinequality}\\
    \leq& \:\gamma\max_{s_t,a_t}\sum_{r_t,s_{t+1}}p(r_t,s_{t+1}|s_t,a_t)\lVert Q^1-Q^2\rVert_{\infty} \\
    \leq& \:\gamma\lVert Q^1-Q^2\rVert_{\infty}\:,\label{equ:first_result}
\end{align}
\else
\begin{align}
    \lVert F&^*Q^1-F^*Q^2\rVert_{\infty} \nonumber\\
    =& \max_{s_t,a_t}\left| (F^*Q^1)(s_t,a_t)-(F^*Q^2)(s_t,a_t)\right| \\
    =& \max_{s_t,a_t}\Big|\sum_{r_t,s_{t+1}}p(r_t,s_{t+1}|s_t,a_t)g(r_t, \gamma \max_{a_{t+1}}Q^1(s_{t+1}, a_{t+1})) \nonumber \\
    &\quad\: -\sum_{r_t,s_{t+1}}p(r_t,s_{t+1}|s_t,a_t)g(r_t, \gamma\max_{a_{t+1}}Q^2(s_{t+1}, a_{t+1}))\Big|  \\
    =& \max_{s_t,a_t} \Big|\sum_{r_t,s_{t+1}}p(r_t,s_{t+1}|s_t,a_t)\big[g(r_t, \gamma \max_{a_{t+1}}Q^1(s_{t+1}, a_{t+1})) \nonumber\\
    &\quad\quad\quad\quad\quad\quad\quad\quad\quad\quad  -g(r_t, \gamma \max_{a_{t+1}}Q^2(s_{t+1}, a_{t+1}))\big]\Big| \\
    \leq& \max_{s_t,a_t}\sum_{r_t,s_{t+1}}p(r_t,s_{t+1}|s_t,a_t)\Big| g(r_t, \gamma \max_{a_{t+1}}Q^1(s_{t+1}, a_{t+1})) \nonumber\\
    &\quad\quad\quad\quad\quad\quad\quad\quad\quad\quad\: -g(r_t, \gamma \max_{a_{t+1}}Q^2(s_{t+1}, a_{t+1}))\Big|\\
    \leq& \: \max_{s_t,a_t}\sum_{r_t,s_{t+1}}p(r_t,s_{t+1}|s_t,a_t)\big|\gamma\max_{a_{t+1}}Q^1(s_{t+1}, a_{t+1}) \nonumber\\
    &\quad\quad\quad\quad\quad\quad\quad\quad\quad\quad\quad\: -\gamma\max_{a_{t+1}}Q^2(s_{t+1}, a_{t+1})\big| \label{equ:use_lemma_1}\\
    \leq& \: \gamma\max_{s_t,a_t}\sum_{r_t,s_{t+1}}p(r_t,s_{t+1}|s_t,a_t)\max_{a_{t+1}}\big| Q^1(s_{t+1}, a_{t+1}) \nonumber\\
    &\quad\quad\quad\quad\quad\quad\quad\quad\quad\quad\quad\quad\quad\quad -Q^2(s_{t+1}, a_{t+1})\big| \label{equ:supinequality}\\
    \leq& \:\gamma\max_{s_t,a_t}\sum_{r_t,s_{t+1}}p(r_t,s_{t+1}|s_t,a_t)\lVert Q^1-Q^2\rVert_{\infty} \\
    \leq& \:\gamma\lVert Q^1-Q^2\rVert_{\infty}\:,\label{equ:first_result}
\end{align}
\fi
where (\ref{equ:use_lemma_1}) follows from the condition \cref{condition_1} in \cref{thm:1}; (\ref{equ:supinequality}) follows from the fact that for any two functions $f_1$ and $f_2$, we have $|\sup_x f_1(x)-\sup_x f_2(x)|\leq \sup_x|f_1(x)-f_2(x)|$; and lastly, (\ref{equ:first_result}) follows from the normalization of the probability distribution $p(r_t,s_{t+1}|s_t,a_t)$.
\end{proof}

With Lemma~\ref{lemma:1} established, we can readily prove the main theorem of the global convergence of the value function updates. 

Starting from any arbitrary value function initialization point $Q^0\in\mathcal{R}^{|\mathcal{V}|}$, consider the value iteration process of iteratively applying the mapping $F^*$. With $\mathcal{R}^{|\mathcal{V}|}$ being a Banach space and $F^*$ being a contraction mapping (by Lemma~\ref{lemma:1}), according to the Banach's fixed-point theorem \cite{banach}, the process is guaranteed to converge to a unique convergence point $Q^*$.  

\section{Proof of \cref{thm:2}}\label{sec:appendix_II}

\begin{lemma}\label{lemma:2}
If $g(\cdot,\cdot)$ satisfies the condition \cref{condition_2} in \cref{thm:2}, then the generalized value function update $F^*$ as in \cref{equ:bellman_update_modified_mapping1} is monotonic, i.e., $\forall\: Q^1, Q^2\in \mathcal{R}^{|\mathcal{V}|}$, if $Q^1\geq Q^2$, then $F^*Q^1\geq F^*Q^2$ always holds\footnote[2]{The notation $Q^1\geq Q^2$ implies $Q^1(s,a)\geq Q^2(s,a), \forall s, a$}.
\end{lemma}
\begin{proof}
\begin{align}
    Q^1\geq &\ Q^2  \nonumber\\
    \implies& Q^1(s,a) \geq Q^2(s,a),\ \forall s,a \\
    \implies& \max_a Q^1(s,a) \geq \max_a Q^2(s,a),\ \forall s \\
    \implies& g(r, \gamma \max_a Q^1(s,a))\geq g(r, \gamma \max_a Q^2(s,a)),\ \forall s,r \label{equ:lemma2_equ1} 
\end{align}
where \cref{equ:lemma2_equ1} follows from the condition \cref{condition_2} in \cref{thm:2}. 

Now we introduce the time step into the equations, and consider any given state and action pair at time $t$: $s_t$ and $a_t$. Let $s=s_{t+1}$ in \cref{equ:lemma2_equ1} be the state the agent is in after executing $a_t$ on $s_t$; and let $r=r_t$ be the reward from executing $a_t$ on $s_t$. We then have:
\ifOneColumn
\begin{align}
    Q^1\geq &\ Q^2 \nonumber \\
    \implies& g(r_t, \gamma \max_a Q^1(s_{t+1},a))\geq g(r_t, \gamma \max_a Q^2(s_{t+1},a)),\ \forall r_t, s_{t+1} \\
    \implies& (F^*Q^1)(s_t,a_t) \geq (F^*Q^2)(s_t,a_t),\ \forall s_t, a_t \\
    \implies& F^*Q^1 \geq F^*Q^2
\end{align}
\else
\begin{align}
    Q^1\geq &\ Q^2 \nonumber \\
    \implies& g(r_t, \gamma \max_a Q^1(s_{t+1},a))\geq g(r_t, \gamma \max_a Q^2(s_{t+1},a)), \nonumber \\
    &\qquad\qquad\qquad\qquad\qquad\qquad\qquad\qquad\qquad \forall r_t, s_{t+1} \\
    \implies& (F^*Q^1)(s_t,a_t) \geq (F^*Q^2)(s_t,a_t),\ \forall s_t, a_t \\
    \implies& F^*Q^1 \geq F^*Q^2
\end{align}
\fi
\end{proof}

To show that $Q^*$ is the optimal point, let $\pi_0$ be an arbitrary initial policy and $Q^0\in\mathcal{R}^{|\mathcal{V}|}$ be the corresponding value function (therefore $Q^0=F^{\pi_0}Q^0$). we have:
\begin{align}
    Q^0 = F^{\pi_0}Q^0 \leq F^*Q^0
\end{align}
where the inequality follows from the maximization over the actions by the greedy policy mapping $F^*$. Applying $F^*$ again on both sides of the inequality, and by the monotonicity from Lemma~\ref{lemma:2} we have:
\begin{align}
    Q^0 \leq F^*Q^0 \leq {F^*}^2Q^0
\end{align}
where $F^2$ denotes iteratively applying the mapping $F$ twice. After iteratively applying $F^*$ until convergence, we arrive at a chain of inequalities ending with the unique convergence point:
\begin{align}
    Q^0 \leq F^*Q^0 \leq {F^*}^2Q^0 \leq {F^*}^3Q^0 \leq \dots
    \leq\lim_{n\to\infty}{F^*}^nQ^*= Q^*
\end{align}
Since $Q^0$ is an arbitrary value function, we have shown that the unique point of convergence $Q^*$ is indeed the global maximum value function.

Furthermore, given the assumptions that $p(r_t|s_t,a_t)$ and $p(s_{t+1}|s_t,a_t)$ are deterministic as required by \cref{thm:2}, we have $g(\cdot,\cdot)$ being exchangeable with $\mathbb{E}_{p(r_t|s_t,a_t)}[\cdot]$ and $\mathbb{E}_{p(s_{t+1}|s_t,a_t)}[\cdot]$. By bringing the expectations inside the operation $g(\cdot,\cdot)$, we can see that \cref{equ:before_exp} and \cref{equ:Q_exp} are equivalent. Correspondingly, the converged point of the value iteration is truly the expectation value of the objective function as defined in \cref{equ:general_obj}. Therefore, the greedy policy $\pi^*$ derived from the value function $Q^*$ is truly the global optimal policy.

\section{Proof of Convergence Properties on the Bottleneck Update Operation}\label{sec:appendix_III}
To show that the bottleneck update operation \cref{equ:Q_min} satisfies the condition by \cref{condition_1} in \cref{thm:1}:
\begin{proof}
    Without loss of generality, we assume $b\geq c$, then we have:
    \begin{itemize}
        \item If $c\leq b\leq a$, then:
        \begin{align}
        |g(a,b)-g(a,c)|=|\min(a,b)-\min(a,c)|=|b-c|.
        \end{align}
        \item If $c\leq a<b$, then:
        \begin{align}
        |g(a,b)-g(a,c)|=&\:|\min(a,b)-\min(a,c)|\\
        =&\ a-c \\
        <&\ |b-c|.
        \end{align}
        \item If $a<c\leq b$, then: \begin{align}
        |g(a,b)-g(a,c)|=&\:|\min(a,b)-\min(a,c)|\\
        =&\ 0\\
        <&\ |b-c|.
        \end{align}
    \end{itemize}
\end{proof}

To show that the bottleneck operation \cref{equ:Q_min} satisfies the condition by \cref{condition_2} in \cref{thm:2}:
\begin{proof}
    Given $b\geq c$, we have:
    \begin{itemize}
        \item If $a\leq c$, then:
        \begin{align}
            g(a,b) = \min(a,b)=a=\min(a,c) = g(a,c).
        \end{align}
        \item If $c< a\leq b$, then:
        \begin{align}
            g(a,b) = \min(a,b)=a> c=\min(a,c) = g(a,c).
        \end{align}
        \item If $a>b$, then:
        \begin{align}
            g(a,b) = \min(a,b)=b\geq c=\min(a,c) = g(a,c).
        \end{align}
    \end{itemize}
\end{proof}

\section{Proof of Convergence Properties on the Harmonic Mean Update Operation}\label{sec:appendix_IV}
To show that the harmonic mean operation \cref{equ:Q_harmonic} satisfies the condition by \cref{condition_1} in \cref{thm:1}:
\begin{proof}
    With our assumption of positive rewards, we have:
    \begin{align}
        |g(a,b)-g(a,c)| =&\: \left|\frac{1}{\frac{1}{a}+\frac{1}{b}}-\frac{1}{\frac{1}{a}+\frac{1}{c}}\right| \\
        =&\: \frac{|b-c|}{(\frac{1}{a}+\frac{1}{b})(\frac{1}{a}+\frac{1}{c})bc} \\
        \leq&\: \frac{|b-c|}{(\frac{1}{b})(\frac{1}{c})bc} \\
        =&\: |b-c| 
    \end{align}
\end{proof}

To show that the harmonic mean operation \cref{equ:Q_harmonic} satisfies the condition by \cref{condition_2} in \cref{thm:2}:
\begin{proof}
    Given $b\geq c$. With our assumption of positive rewards, we have:
    \begin{align}
        g(a,b)-g(a,c) =&\: \frac{1}{\frac{1}{a}+\frac{1}{b}}-\frac{1}{\frac{1}{a}+\frac{1}{c}}\\
        =&\:\frac{b-c}{(\frac{1}{a}+\frac{1}{b})(\frac{1}{a}+\frac{1}{c})bc} \\
        \geq&\: 0 
    \end{align}
\end{proof}

\bibliographystyle{IEEEtran}
\bibliography{IEEEabrv, References}

\end{document}